\documentclass[10pt,journal,compsoc]{IEEEtran}
\usepackage{graphicx}
\usepackage{subcaption}
\usepackage{amsmath}
\usepackage{amsthm}
\usepackage{bm}
\usepackage[linesnumbered,boxed,ruled,commentsnumbered]{algorithm2e} 
\usepackage{amssymb}
\usepackage{lineno,hyperref}
\usepackage{latexsym}
\usepackage{ragged2e}
\ifCLASSOPTIONcompsoc
  \usepackage[nocompress]{cite}
\else
  \usepackage{cite}
\fi
\ifCLASSINFOpdf

\else

\fi

\usepackage{framed,multirow}

\usepackage{tabularx, environ}

\usepackage{tikz}
\usepackage{balance}

\newtheorem{problem}{Problem}
\newtheorem{definition}{Definition}

\begin{document}

\title{Variational Transfer Learning using Cross-Domain Latent Modulation}

%

%
%

\author
{Jinyong~Hou, Jeremiah~D.~Deng,~\IEEEmembership{Senior Member,~IEEE,} 
Xuejie~Ding, 
and~Stephen~Cranefield%
\thanks{J. Hou, J. D. Deng, X. Ding and S. Cranefield are with the 
School of Computing, University of Otago, Dunedin 9054, New Zealand. E-mail:
{jeremiah.deng@otago.ac.nz}
}}

\IEEEpeerreviewmaketitle
\IEEEtitleabstractindextext{
\begin{justify}
\begin{abstract}
    For successful application of neural network models to new domains, powerful transfer learning solutions are essential.
    We propose to introduce a novel cross-domain latent modulation mechanism to a variational autoencoder framework so as to achieve effective transfer learning for tasks such as domain adaptation and image translation. Our key idea is to extract deep representations from one data domain and use it to drive the reparameterization of the latent variable of another domain. Specifically, deep representations of the source and target domains are first extracted by a unified inference model and aligned by employing gradient reversal. The learned deep representations are then cross-modulated to the latent encoding of the alternative domain, where consistency constraints are also applied. 
    Our empirical validations 
    demonstrate the competitive performance of the proposed framework, which is also supported by evidence obtained from visualization and qualitative evaluation. 
   
\end{abstract}

\end{justify}
}

\maketitle

\IEEEdisplaynontitleabstractindextext
\IEEEpeerreviewmaketitle


\section{Introduction} \label{sec:introduction}

In machine learning, the out-of-sample error is not the only factor that challenges the generalization ability of a  model. In addition to deal with new data for testing, we may wish to apply our model to new domains, or new tasks.  However, the performance of the trained model often plunges significantly, as the data acquired from the new domain may have significant sampling bias, or simply come from a very different probability distribution. To tackle this issue, transfer learning has been intensively studied in recent years, which aims to adapt the knowledge obtained from the current domain (the ``source'') to a related domain or task (the ``target'')~\cite{Pan2010, Tan2018, Weiss2016}. Very often, domain adaptation can be done in an unsupervised manner, i.e., without requiring the availability of labels from the target domain. 

From the perspective of probabilistic modelling~\cite{Liu2017, Wang2017a, Schonfeld2019}, to achieve cross-domain transfer, it is crucial to learn a joint distribution of data from different domains. Once the joint distribution is learned, it can generate the marginal distributions of the individual domains~\cite{Kingma2013a, Liu2017}. Using the variational inference approach, an inferred joint distribution is often applied to the variational latent space. 
However, inferring the joint distribution from the marginal distributions of different domains is a highly ill-posed problem~\cite{Hollander2012}. To address this problem, UNIT~\cite{Liu2017} assumes a shared latent space for the source and target domains. This can be achieved by applying an adversarial learning strategy to the domains' latent spaces. Another approach is to use a complex prior to improve the input data's representation performance~\cite{Mahajan2020, Tomczak2018, Hoffman2016}. As far as we know, there are few works that have examined the potential cross-domain generation processes within the latent space, which we argue could help cross-domain transfer scenarios.  

In this paper, we present a novel latent space reparameterization framework that employs a generative, cross-domain latent modulation process to cater for the cross-domain transferability. Specifically, we first propose a transfer latent space (TLS) to build a general cross-domain latent space. We then incorporate cross-domain modulation mechanisms into the reparameterization process, which generates transfer latent representations with reduced Kullback-Leibler divergence. The generated transfer latent space is further tuned by domain-level adversarial alignment and inter-class alignment, applying a cross-consistency loss to decoded data obtained through reconstruction and generations. We call our model ``Cross-Domain Latent Modulation'' (CDLM). The latent modulation could be regarded as a specific instance of the proposed TLS. In this paper, we explore two kinds of cross-domain modulation operations. 
For experiments, the CDLM model is applied to the homogeneous transfer scenarios including unsupervised domain adaptation and image-to-image translation, where highly competitive performance is reported. Further supportive evidence obtained from ablation studies and visualization is also presented. 

The CDLM framework was first presented in our previous work~\cite{Hou2021}, where some preliminary results were reported. In this paper, we present more theoretical analyses and empirical studies of the framework. Specifically, the following new contributions are made: 1) We extend the proposed framework to allow for both complete and partial modulations; 2) In addition to giving a geometric interpretation of the CDLM mechanism, we employ the Kullback-Leibler divergence to demonstrate the reduction of domain gap resulted from the latent modulations; 3) 
Finally, for the empirical study we have also included more state of the art methods for comparison, and have new performance results and visualization outcome reported. 

The rest of the paper is organized as follows. In Section~\ref{sec:related_work}, some related work is briefly reviewed. In Section~\ref{sec:model}, we outline the overall structure of our proposed model, carry out analysis on K-L divergence, and develop learning metrics with the defined losses. The experiments are presented and discussed in Section~\ref{sec:experiments}. We conclude our work in Section~\ref{sec:conclusion}, and point to some future work.

\section{Related Work} \label{sec:related_work}

\begin{figure}[!thbp]
  \centering
  \includegraphics[width=0.6\columnwidth]{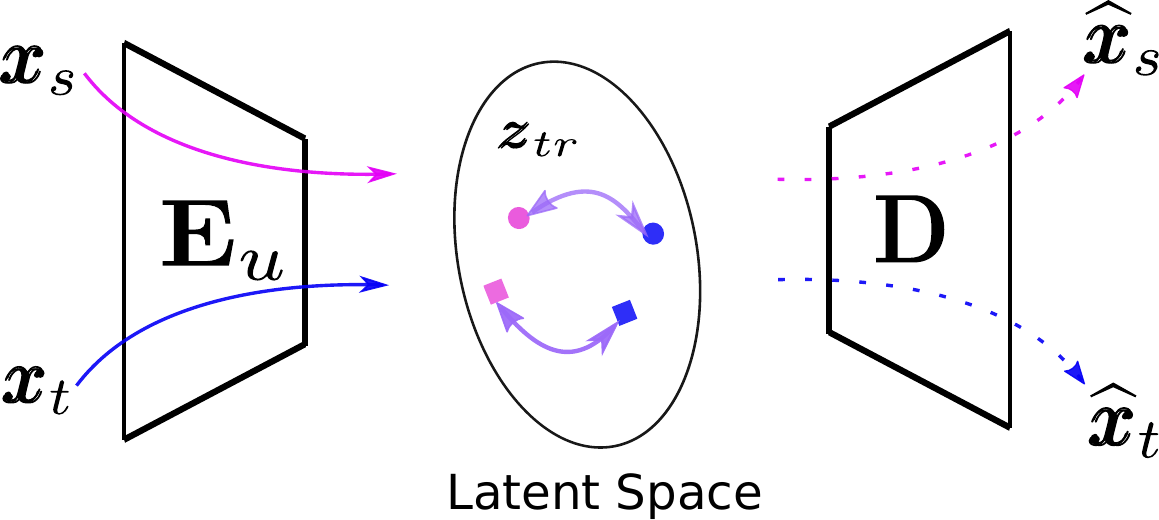}
  \caption[\small Illustration of the latent space manipulation.]{\small Illustration of the latent space manipulation. We can get the latent space by a deep autoencoder structure ($\mathbf{E}_u$ stands for an encoder and $\mathbf{D}$ for a decoder). Our aim is then to design a transformation to transfer the latent space ($\bm z_{tr}$) for the source and the target. The source data flow ($\bm x_s, \bm \hat{x}_s$) is in purple and the target ($\bm x_t, \bm \hat{x}_t$) is in blue.}
  \label{fig:ch4_schematic_latent_space_manipulation}
\end{figure}

To model a joint distribution for cross-domain adaptation, manipulation of latent spaces is a common approach~\cite{Larsen2016, Liu2017, Liu2018b}. A shared latent space for the source and target domains is employed, as illustrated in~Fig.~\ref{fig:ch4_schematic_latent_space_manipulation}, which provides latent encodings as common representations for data inputs across domains. Some adversarial strategy is usually used to pull the latent representations closer so that they are less domain-dependent.

Under the variational inference context, we assume a prior for the estimated latent space. It is easy to confine the joint distribution into the same space and apply the adversarial strategy. For example, works~\cite{Liu2017, Schonfeld2019, Liu2018b} assume a standard Gaussian prior for the transfer. Also, we can further manipulate the variational latent space by the priors. Works in~\cite{Mahajan2020, Ziegler2019, Kingma2016} adopt normalizing flows as complex priors for multimodal latent representations. 

In~\cite{Kingma2016}, inverse autoregressive flow (IAF) is proposed for the variational inference, which scales well to high-dimensional latent spaces. IAF is constructed by the successive transformations based on the autoregressive neural network. It improves upon diagonal Gaussian approximation posterior significantly. Reference~\cite{Ziegler2019} introduces the normalizing flow for discrete sequences. It utilizes a VAE-based generative model that jointly learns a normalizing flow distribution in the latent space and a stochastic mapping to an observed discrete space. To this end, it is important that the flow-based distribution be highly multi-modal. Another interesting application is for cross-domain many-to-many mapping described in~\cite{Mahajan2020}. It learns shared joint representations between images and text, also domain-specific information separately. The normalizing flow-based prior is used for the domain-specific information to learn various many-to-many mappings. 

Gaussian mixture~\cite{Bishop2006Pattern} is also employed as prior of the variational latent space for a better cluster performance. Reference~\cite{Dilokthanakul2016} proposes a variant of variational autoencoders (VAEs) with a Gaussian mixture prior to improve the unsupervised clustering performance. In~\cite{Wang2017a}, conditional variational autoencoders~\cite{Sohn2015} (CVAEs) is utilized with a Gaussian mixture and a novel additive Gaussian prior to create a prior for images that contain different content simultaneously. It can enhance the image caption generation's variability. Reference~\cite{Yang2019b} improves the deep clustering using Gaussian mixture variational autoencoders(VAEs) with graph embedding. The Gaussian mixture model (GMM) is used for the local multi-clustering latent space, and the graph structure is for the global information. The combination can learn powerful representations for the inputs. 

Another approach is to use disentangled latent representations where the latent encoding is divided into some defined parts (e.g.~style and content parts), then the model learns separated representations and swaps them for the transfer~\cite{Bousmalis2016, Zhang2019d, Feng2018a, Lee2018, Gonzalez-garcia2018}. For example, in~\cite{Lee2018}, images are mapped into a domain-invariant content space and a domain-specific attribute space for each domain separately. Different encoders are employed for the source and target domains, respectively. Each domain has its decoder. However, the latent encoding is done in a cross-domain manner by swapping the learned content between domains. The reconstructions with swapped content are re-encoded in a cross-domain manner for cycle-consistency. During the image mapping, an adversarial strategy is utilized for the unsupervised alignment. 

Reference~\cite{Bousmalis2016} has a similar strategy. It splits the encodings into similar and different parts for each domain, extracted by two distinguishing encoders. The combination of the learned domain-invariant and domain-private information is fed into a shared decoder for the reconstructions. The model uses a squared Frobenius norm as a difference regularization, adversarial strategy, and MMD for the shared information. In~\cite{Gonzalez-garcia2018}, a novel network structure is proposed with a generative adversarial network (GANs) and cross-domain autoencoder. It aims to separate the internal representation into three parts: shared and exclusive parts for domains by paralleled variational autoencoders.

The manipulation on the latent space is often interwoven with the homogeneous image transfer together, such as unsupervised domain adaptation and image translation~\cite{Naseer2019,Noguchi2019,Choi2018}. In the domain separation networks~\cite{Bousmalis2016}, separate encoding modules are employed to extract the invariant representation and domain-specific representations from the domains respectively, with the domain-invariant representations being used for the domain adaptation. References~\cite{Bousmalis2017,Sankaranarayanan2018,Hoffman2018} transfer the target images into source-like images for domain adaptation.  Works in~\cite{Liu2017,Zhu2017,Kim2017} map the inputs of different domains to a single shared latent space, but require the cycle consistency for the completeness of the latent space. The UFDN~\cite{Liu2018b} utilizes a unified encoder to extract the multi-domain images to a shared latent space, and the latent domain-invariant coding is manipulated for image translation between different domains. 

Our method is different from these approaches. In our model, a learned deep auxiliary representation is used to generate perturbations to the latent space through a modified reparameterization trick using variational information from the counterpart domain. It helps generate cross-domain image translation. The transfer is carried out by a reparameterization transformation, using statistical moments retaining specific information for one domain, and deep representation providing information from the other. Our model adopts the pixel-level adaptation between domains from the cross-domain generation, but the proposed model can also be used at the feature-level due to the latent space alignment. Our model also has a unified inference model, but the consistency is imposed straightforwardly, reducing computational complexity. 


\section{The Proposed Model} \label{sec:model}

We now present our proposed CDML model in detail. First, we formally describe the problem settings. The proposed transfer latent space (TLS) and modulation mechanism are then described. Following this, we give two specific implementations of cross-domain modulation in the TLS, and demonstrate that they contribute to the reduction of Kullback-Leibler divergence between TLS representations. Finally, our model's learning process is presented. 

We first give a summary of all the notations we frequently use from now on. 
\begin{table}[h]
   \centering
   \caption{\small List of notations used in the description of CDLM.}
     \begin{tabular}[t]{c c}
       \hline 
       Notation & Description \\
       \hline
       $\mathbf{X}_s/\bm y_s$ & Source data and labels \\  
       $\mathbf{X}_t/\bm y_t$ & Target data and labels \\
       $\mathbf{E}_{\bm \phi}$ & Encoder \\
       $\mathbf{D}_{\bm \theta}$ & Decoder \\ 
       $\bm h_s$ & Source's high-level representation \\ 
       $\bm h_t$ & Target's high-level representation  \\
       $\ddot{\bm z}_{s}$ & Source's latent encoding  \\ 
       $\ddot{\bm z}_{t}$ & Target's latent encoding  \\ 
       $\ddot{\bm z}_{st}$ & Source's modulated latent encoding  \\ 
       $\ddot{\bm z}_{ts}$ & Target's modulated latent encoding  \\ 
       $\Xi$ & Discriminator \\ 
       $\mathbf{W}_{\mu}/\bm b_{\mu}$ & Weights / bias for latent mean $\bm\mu$ \\ 
       $\mathbf{W}_{\sigma}/\bm b_{\sigma}$ & Weights / bias for latent standard deviation $\bm\sigma$ \\ 
       $\mathbf{W}_{\bm h}/\bm b_{\bm h}$ & Weights / bias for high-level representations \\
       \hline 
    \end{tabular}
   \label{table:ch4_notations_cdlm}
\end{table}

\begin{problem}{}
  Let $\mathcal{X} \subset \mathbb{R}^d$ be a $d$-dimensional data space, and $\mathbf{X} = \{\bm x_1, \bm x_2, \ldots, \bm x_n\} \in \mathcal{X}$ the sample set with marginal distribution $p(\mathbf{X})$. The source domain is denoted by a tuple $(\mathcal{X}_s, p(\mathbf{X}_s))$, and the target domain by $(\mathcal{X}_t, p(\mathbf{X}_t))$. We consider the so-called homogeneous transfer with domain shift, i.e. $\mathcal{X}_s \approx \mathcal{X}_t$, but $p(\mathbf{X}_s) \ne p(\mathbf{X}_t)$. 
  A label set is $\mathbf{y}=\{y_1, y_2, \ldots y_n\} \in \mathcal{Y}$ ($\mathcal{Y}$ is the label space), and only the source domain's label set $\mathbf{Y}_s$ is available during transfer learning. Therefore, our learning task is $\mathcal{T} = p(\mathbf{y}_t|\mathbf{X}_s, \mathbf{y}_s; \mathbf{X}_t)$. 
\end{problem}

\subsection{Transfer latent space}
\label{subsec:ch4_transfer_latent_space}

As an infinite number of joint distributions can yield any given marginal distribution, we need to construct an inference framework under some constraints. Our approach is to use the latent space within the variational autoencoder (VAE) framework as the manipulation target. We define the concept of ``transfer latent space'' (TLS) as follows.

\begin{definition}
Transfer Latent Space. 
  Let $\bm x_s \in \mathbf{X}_s$, $\bm x_t \in \mathbf{X}_t$ be the domain samples. Let us have a map $f$ that extracts domain information $\bm\Omega$ and a deep feature representation $\bm h$ from a given input $\bm x$:
  \[ f \colon \bm x \longrightarrow (\bm\Omega, \bm h), \bm x \in \mathbf{X}_s \cup \mathbf{X}_t.\] 
  Suppose we construct a transfer map $\mathcal{G}$ that generates the following latent variables 
  from $\bm\Omega$ and $\bm h$ with domain crossovers:
  \[ \bm{\ddot{z}}_{st}=\mathcal{G}(\bm\Omega_s,\bm h_t), \]
  \[ \bm{\ddot{z}}_{ts}=\mathcal{G}(\bm\Omega_t,\bm h_s). \]
  The joint space formed by $\bm{\ddot{z}}_{st}$ and $\bm{\ddot{z}}_{ts}$ samples is defined as the \textit{transfer latent space} (TLS), denoted by $\ddot{\bm Z}$. 
\end{definition}

As can be seen, the TLS is intended to become a ``mixer'' for the two domains, because the resulted latent variables are under cross-domain influences. Therefore, our transfer latent space can be regarded as a generalization of the latent space under cross-domain settings.

Our definition of the TLS finds an analogy in telecommunications, where data are first modulated by a carrier signal so that it can propagate through a communication channel. For example, in the case of frequency modulation, the frequency of the transmission signal is composed by the carrier frequency $f_c$ and the data bit $b$: \[ f = f_c + b\Delta f, \] where $\Delta f$ is the frequency deviation. 

Similarly, the TLS builds a cross-domain latent space by a ``modulation'' between the $\Omega_s$ and ${\bm h}_t$, and vice versa. Therefore, we call our model ``Cross-Domain Latent Modulation'', or CDLM for short. 

\subsection{The CDLM framework}
\label{subsec:ch4_cdlm_framework}
\subsubsection{Overview}
\begin{figure*}[!thbp]
  \centering
  \includegraphics[width=0.7\textwidth]{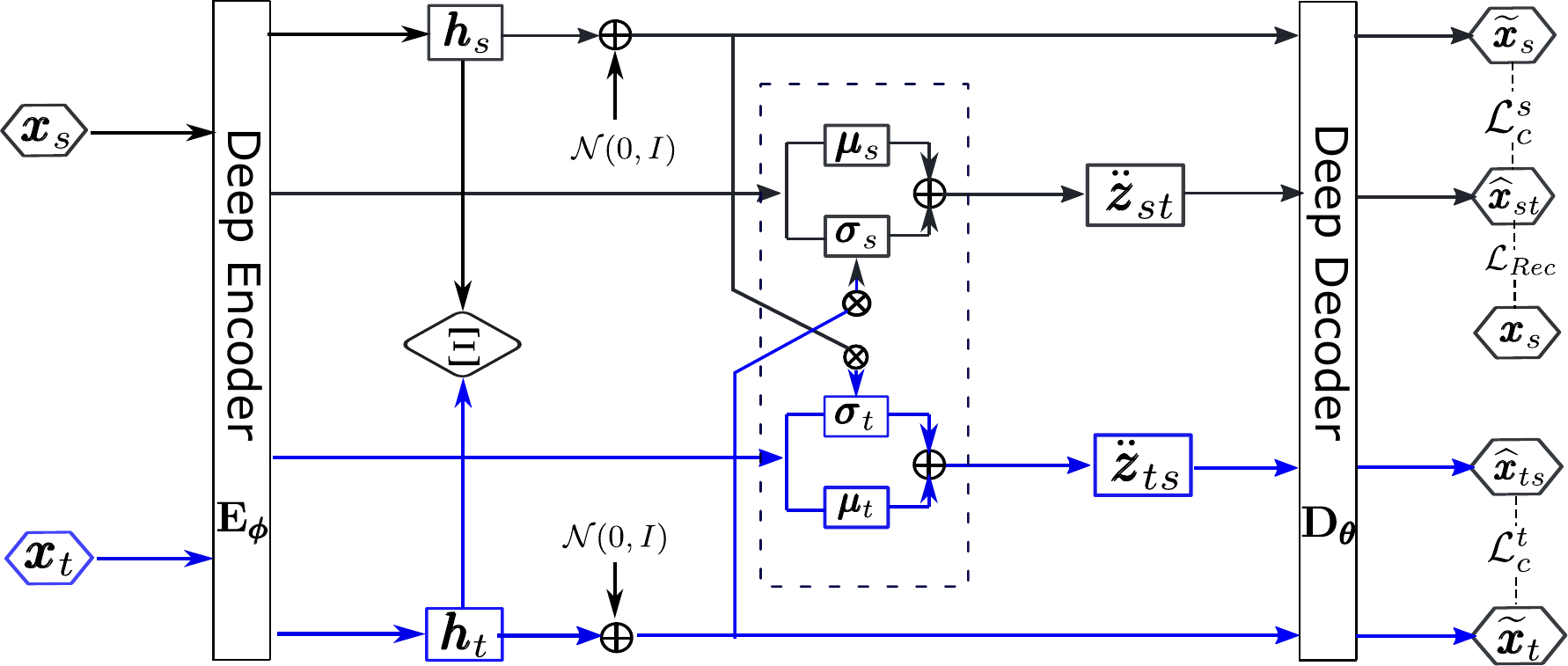}
  \caption[\small Architectural view of the proposed CDLM.]{\small Architectural view of the proposed model~\cite{Hou2021}. It encourages an image from the target domain (blue hexagon) to be transformed to a corresponding image in the source domain (black hexagon). The transfer latent distributions $p(\bm {\ddot{z}}_{ts}|\bm x_t, \bm h_s)$ and $p(\bm{\ddot{z}}_{st}|\bm x_s, \bm h_t)$ are learned and are used to generate corresponding images by the desired decoder. The deep representations are integrated into the reparameterization transformation with standard Gaussian auxiliary noise. Blue lines are for the target domain and black ones are for the source domain.}
  \label{fig:model_sche}
\end{figure*}

Our proposed CDLM framework is shown in Fig.~\ref{fig:model_sche}. In the framework, we build the cross-domain generation by a unified inference model $\mathbf{E}_{\bm \phi}(\cdot)$ (as an implementation of the $f$ map) and a generative model for the desired domain $\mathbf{D}_{\bm \theta}(\cdot)$, e.g., the source domain in our model. A discriminator $\Xi$ is utilized for the adversarial training. We use the terms ``inference model'' and ``encoder'' for $\mathbf{E}_{\bm \phi}(\cdot)$, and ``generative model'' and ``decoder'' for  $\mathbf{D}_{\bm \theta}(\cdot)$ interchangeably.

As discussed in Section~\ref{subsec:ch4_transfer_latent_space}, under the variational framework, the domain information $\bm \Omega$ (here we remove the domain subscript for simplicity) is usually the Gaussian distribution parameters pair $(\bm \mu, \bm \sigma)$. Let $\bm h'$ be the flattened activations of the last convolution layer in $\mathbf{E}_{\bm \phi}$. Then, following the treatment in~\cite{Kingma2014}, $\bm \mu$ and $\bm \sigma$ can be obtained by $\bm \mu = \mathbf{W}_{\bm \mu}\bm h' + \bm b_{\bm \mu}$ and $\log\bm \sigma = \mathbf{W}_{\bm \sigma}\bm h' + \bm b_{\bm \sigma}$, where $\mathbf{W}_{\bm \mu}, \mathbf{W}_{\bm \sigma}, \bm b_{\bm \mu}, \bm b_{\bm \sigma}$ are the weights and biases for $\bm \mu$ and $\bm \sigma$.

\begin{figure}[h]
  \centering
  \includegraphics[width=0.5\columnwidth]{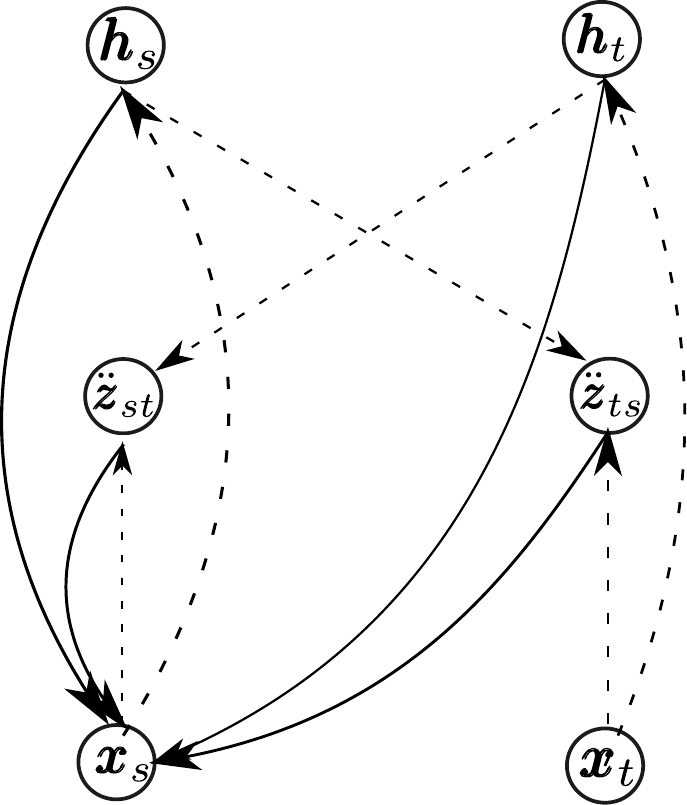}
  \caption[\small Graphical model of CDLM.]{\small Graphical model of CDLM. The dashed lines are inference process and the solid lines are generative process.}
  \label{fig:graphical_model_cdlm}
\end{figure}

From our observations, both shallow (e.g., PCA features) and deep representations can be used to obtain domain information $\bm h$. In our end-to-end model we use the latter. We choose the activations of the last convolutional layer, i.e., $\bm h = \text{sigmoid}(\mathbf{W}_{\bm h}\bm h' + \bm b_{\bm h})$ as the high-level representation~\cite{Yosinski2014}, where $\mathbf{W}_{\bm h}, \bm b_{\bm h}$ are the weights and biases for the deep abstractions. 
Having obtained the domain information $\bm\Omega$ and deep representation $\bm h$, a natural choice for the transfer map $\mathcal{G}$ is through reparameterization. We now propose a modified reparameterization trick to sample from 
the transfer latent space, as follows:
\begin{equation}  \label{eq:zst_cdlm_latent}
  \bm{\ddot{z}}_{st} = \mathcal{G}((\bm\mu_s, \bm\sigma_s), \bm h_t) = \bm\mu_s + \bm\sigma_s \odot (\gamma_1\bm h_t + \gamma_2\bm\epsilon),
\end{equation}
and
\begin{equation}  \label{eq:zts_cdlm_latent}
  \bm{\ddot{z}}_{ts} = \mathcal{G}((\bm\mu_t, \bm\sigma_t), \bm h_s) = \bm\mu_t + \bm\sigma_t \odot (\gamma_1\bm h_s + \gamma_2\bm\epsilon),
\end{equation}
where $\bm h_s$ ($\bm h_t$) is the sample of the deep representation space $\mathcal{H}_s$ ($\mathcal{H}_t$); $\bm \mu_s$ and $\bm\sigma_s$ ($\bm \mu_t$ and $\bm\sigma_t$) are the mean and standard deviation of the approximate posterior for the source (target) domain; $\gamma_1, \gamma_2>0$ are trade-off hyperparameters to weight the deep feature modulation and the standard Gaussian noise $\bm\epsilon$; and $\odot$ stands for the the Hadamard product of vectors. Therefore, the auxiliary noise in VAEs resampling is now a weighted sum of a deep representation from the other domain and Gaussian noise, different from the standard VAEs framework. The graphical model of CDLM is shown in Fig.~\ref{fig:graphical_model_cdlm}.

\subsubsection{Full CDLM and divergence analysis}
\label{sec:KLDana}
Specifically, we take a closer look of the modulated reparameterization. Assume $\bm h_t \sim p(\mathcal{H}_t) = \mathcal{N}(\bm \mu_{\bm h_t}, \bm \sigma_{\bm h_t})$. Then for $\bm{\ddot{z}}_{st}$, we have $\bm{\ddot{z}}_{st} \sim \mathcal{N}(\bm{\ddot{z}}_{st}; \bm \mu_{st}, \bm \sigma_{st}^2\mathbf{I})$. For the $i$-th element of the distribution mean vector is given as follows:
\begin{equation}  \label{eq:ch4_new_mu}
  \begin{split}
    \mu_{st}^i &= \mathbb{E}\left[\mu^i_s + \sigma_s^i(\gamma_1 h_t^i + \gamma_2 \epsilon^i)\right ]\\
    & = \mu_s^i + \sigma_s^i\gamma_1 E\left[h_t\right] + \gamma_2 E\left[\epsilon^i\right] \\
    &= \mu_s^i + \gamma_1 \sigma_s^i \mu_{h_t}^i,
  \end{split}
\end{equation}
and the $i$-th variation element is obtained as
\begin{equation}  \label{eq:ch4_new_sigma}
  \begin{split}
    \mathbf{Var}(\ddot{z}_{st}^i) &= \mathbf{Var}[\sigma_s^i (\gamma_1 h_t^i + \gamma_2 \epsilon^i)]\\
    &=\mathbb{E}\left[(\sigma_s^i(\gamma_1 h_t^i + \gamma_2\epsilon^i) - \gamma_1\sigma_s^i\mu_{h_t}^i)^2\right] \\
    &= (\sigma_s^i)^2 \mathbb{E}\left[( (\gamma_1 h_t^i + \gamma_2\epsilon^i) - \gamma_1\mu_{h_t}^i)^2\right] \\
    &= (\sigma_s^i)^2 \left( \mathbb{E}\left[( (\gamma_1 h_t^i + \gamma_2\epsilon^i))^2\right] - (\gamma_1 \mu_{h_t}^i)^2 \right) \\
    &= (\sigma_s^i)^2 [\gamma_1^2(\sigma_{h_t}^i)^2 + \gamma_2^2].
  \end{split}
\end{equation}

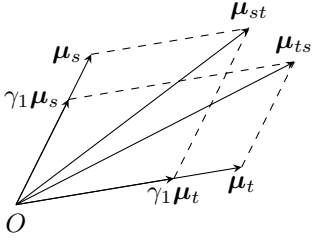
\begin{figure}[!thbp] 
  \centering
\begin{tikzpicture}
\draw (0,0) node[below] {$O$};
\draw[-stealth] (0,0)--(3,0.5);
\draw (3,0.5) node[below]{$\bm\mu_t$};
\draw[-stealth] (0,0)--(2.1,0.35);
\draw (2.1,0.35) node[below]{$\gamma_1\bm\mu_t$};
\draw[-stealth] (0,0)--(1,2);
\draw (1,2) node[left]{$\bm\mu_s$};
\draw[-stealth] (0,0)--(.7,1.4);
\draw (.7,1.4) node[left]{$\gamma_1\bm\mu_s$};
\draw[dashed] (2.1,0.35)--(3.1,2.35);
\draw[dashed] (1,2)--(3.1,2.35);
\draw[-stealth] (0,0)--(3.1,2.35);
\draw (3.1,2.35) node[above]{$\bm\mu_{st}$};
\draw[dashed] (.7,1.4)--(3.7,1.9);
\draw[dashed] (3,.5)--(3.7,1.9);
\draw[-stealth] (0,0)--(3.7,1.9);
\draw (3.7,1.9) node[above]{$\bm\mu_{ts}$};
\end{tikzpicture}
  \caption{Modulation effect on the latent mean vectors: transfer means $\bm\mu_{st}$ and $\bm\mu_{ts}$ are pushed closer compared with the original domain latent means $\bm\mu_{s}$ and $\bm\mu_{t}$.}
  \label{fig:latent_mu_analysis_geometry}
\end{figure}

Here, it is reasonable to assume  $\bm \mu_{\bm h_t} \approx \bm \mu_t$ and $\bm \sigma_{\bm h_t} \approx \bm \sigma_t$ with sufficient training. With a practical setting of $\gamma_1\gg\gamma_2$, and in effect $\sigma_s^i\approx 1$, Eqs.~(\ref{eq:ch4_new_mu} and \ref{eq:ch4_new_sigma}) can be further simplified to 
\begin{equation}
    \mu_{st}^i=\mu_s^i+\gamma_1\mu_t^i,
\label{eq:mu_st}
\end{equation}
and 
\begin{equation}
\sigma_{st}^i=\gamma_1\sigma_s^i\sigma_t^i.    
\label{eq:sigma_st}
\end{equation}
Then we can see that the effect of cross modulation $\bm{\ddot{z}}_{st}$ is to apply a shift to the mean of the original latent mean $\bm\mu_s$, under the influence of $\bm \mu_t$, which results in the new cross-modulated latent mean $\bm \mu_{st}$; and $\bm\sigma_{st}$ can be taken as a recolouring of $\bm\sigma_s$ under the influence from the target domain. The formulation of $\bm{\ddot{z}}_{ts}$ can be similarly worked out and interpreted. These modulated encodings are hence constructed in a cross-domain manner. The geometric interpretation of the generation of $\bm \mu_{st}$ from $\bm \mu_{s}$ and $\bm \mu_{t}$, as given by Eq.(\ref{eq:mu_st}), is illuatrated in Fig.~\ref{fig:latent_mu_analysis_geometry}, which indicates reduced domain gap between the two cross-modulations than between their original domain counterparts.


We now formally show that the cross modulations results in reduced KL-divergence. 
For the original domains, the KL-divergence between them can be formulated by: 
\begin{equation}
    D_\mathrm{KL}(s||t)=\frac{1}{2}\sum_i\left[2\log\frac{\sigma_t^i}{\sigma_s^i} + \left(\frac{\mu_s^i-\mu_t^i}{\sigma_t^i}\right)^2+\left(\frac{\sigma_s^i}{\sigma_t^i}\right)^2-1\right],
\label{eq:KLD_s_t}
\end{equation}
which can be further simplified into 
\begin{equation}
    D_\mathrm{KL}(s||t)=\frac{1}{2}\sum_i\left[ \left(\frac{\mu_s^i-\mu_t^i}{\sigma_t^i}\right)^2\right].
\label{eq:KLD_s_t_simple}
\end{equation}
when we consider the training effect on source and target latent variables that makes $\sigma_s^i=\sigma_t^i=1$. 

Similar to Eqs.(\ref{eq:mu_st}) and (\ref{eq:sigma_st}), we can also obtain $\mu_{ts}^i=\mu_t^i+\gamma_1\mu_s^i$, and $\sigma_{ts}^i=\gamma_1\sigma_s^i\sigma_t^i=\sigma_{st}^i$. Hence, adapting Eq.(\ref{eq:KLD_s_t}) for the transfer label domains, we have 
\begin{equation}
    D_\mathrm{KL}(st||ts)=\frac{1}{2}\sum_i\frac{(\mu_{st}^i-\mu_{ts}^i)^2}{(\gamma_1\sigma_t^i)^2}=\frac{1}{2}\frac{\left[(1-\gamma_1)(\mu_s^i-\mu_t^i)\right]^2}{(\gamma_1\sigma_t^i)^2},
\end{equation}
which suggests that 
\begin{equation}
    D_\mathrm{KL}(st||ts)=\left(\frac{1-\gamma_1}{\gamma_1}\right)^2 D_\mathrm{KL}(s|t).
\end{equation}
Clearly, with $\gamma_1>0.5$, we have $D_\mathrm{KL}(st||ts)<D_\mathrm{KL}(s|t)$. Likewise, $D_\mathrm{KL}(ts||st)<D_\mathrm{KL}(t|s)$. Therefore, the domain gap between the original domains ``s'' and ``t'' is reduced through their transferred versions ``$st$'' and ``$ts$''.  
\subsubsection{Partial modulation and analysis}
\label{sec:partial-KLD}
We now introduce a couple of variants of latent modulation $\mathcal{G}(\cdot)$ in the transfer latent space (TLS).
Instead of employing the deep representations for modulation as in Eqs.(\ref{eq:zst_cdlm_latent}) and (\ref{eq:zts_cdlm_latent}), we can simply use a mix of latent variances, as follows: 

\begin{equation}  \label{eq:partial_latent1}
\left\{ 
\begin{split} 
& \bm{\ddot{z}}_{st} = \mathcal{G}((\bm\mu_s, \bm \sigma_s, \bm\sigma_t), \bm h_t) = \bm\mu_s + \frac{1}{2}(\bm \sigma_s + \bm\sigma_t) \odot \bm\epsilon; \\
&  \bm{\ddot{z}}_{ts} = \mathcal{G}((\bm\mu_t, \bm \sigma_t, \bm\sigma_s), \bm h_s) = \bm\mu_t + \frac{1}{2}(\bm\sigma_s + \bm \sigma_t) \odot \bm\epsilon 
\end{split}\right.
\end{equation}


The latent modulations in Eq.~(\ref{eq:partial_latent1}) only use the second-order moments for modulation, hence we  call them \textit{partial modulations}, denoted by ``PM1''.  

We can look at the effect of PM1 on KD-divergence. As can be seen, $\bm\mu_{st}=\bm\mu_{s}$, and $\bm\mu_{ts}=\bm\mu_{t}$; also $\bm\sigma_{st}=\bm\sigma_{ts}=\frac{1}{2}(\bm\sigma_s + \bm \sigma_t)$. We have 
\begin{equation}
    D_\mathrm{KL}(st||ts)=D_\mathrm{KL}(ts||st)=\frac{1}{2}\sum_i\left[4\left(\frac{\mu_{s}^i-\mu_t^i}{\sigma_s^i+\sigma_t^i}\right)^2\right].
\end{equation}
We can compare this with Eq.(\ref{eq:KLD_s_t_simple}
). Since 
\[ \frac{4}{(\sigma_s^i+\sigma_t^i)^2}\leq \frac{1}{\sigma_s^i\sigma_t^i} \leq\frac{(\sigma_s^i)^2+(\sigma_t^i)^2}{2(\sigma_s^i\sigma_t^i)^2} =\frac{1}{2}\left(\frac{1}{(\sigma_s^i)^2} + \frac{1}{(\sigma_t^i)^2}\right), \]
and 
\begin{equation}
\begin{split}
   D_\mathrm{KL}(s||t)+D_\mathrm{KL}(t||s) & =\frac{1}{2}\left(\log\frac{\sigma_t^i}{\sigma_s^i}+\log\frac{\sigma_s^i}{\sigma_t^i}\right) \\
    & +\frac{1}{2}\left[ (\mu_s^i-\mu_t^i)^2\left(\frac{1}{(\sigma_s^i)^2}+\frac{1}{(\sigma_t^i)^2}\right) \right] \\
    & + \frac{1}{2}\left[\left(\frac{\sigma_s^i}{\sigma_t^i}\right)^2 + \left(\frac{\sigma_t^i}{\sigma_s^i}\right)^2 -2\right] \\
    & \ge \frac{1}{2}\left[ (\mu_s^i-\mu_t^i)^2\left(\frac{1}{(\sigma_s^i)^2}+\frac{1}{(\sigma_t^i)^2}\right) \right]
\end{split}
\end{equation}
we have 
\begin{equation}
    D_\mathrm{KL}(st||ts)+D_\mathrm{KL}(ts||st)\leq D_\mathrm{KL}(s||t)+D_\mathrm{KL}(t||s),
\end{equation}
i.e., the partial modulation also helps to reduce the domain gap in the transfer latent space. 

We can consider a simplified version of PM1, involving only the standard deviations from the \textit{other} domain, as follows:
\begin{equation}
 \label{eq:partial_latent2}
\left\{ 
\begin{split} 
& \bm{\ddot{z}}_{st} = \bm\mu_s + \bm\sigma_t \odot \bm\epsilon; \\
&  \bm{\ddot{z}}_{ts} = \bm\mu_t + \bm \sigma_s \odot \bm\epsilon 
\end{split}\right.
\end{equation}
It can be shown that this modulation does not reduce the domain gap but preserves it -- further operations are needed to align the domain modulations. We denote this simple partial modulation scheme by ``PM2''. 

\subsubsection{More components}
The transfer encodings are obtained by a shared encoder, which confines the cross-modulated latent variables into the same latent space. 
We have also shown that both full and partial modulations work to reduce the KL-divergence between domains in their transferred latent representations. 
%
However, there is no guarantee yet for the latent representations to be aligned. 
To pull the domains even closer, an adversarial strategy~\cite{Ganin2016, Tzeng2017} is employed to facilitate the alignment. A gradient reversal layer~\cite{Ganin2016} is used in our model, by which adversarial learning is carried out to learn transferable features that are robust to domain shifts. 


Finally, a desired marginalized decoder, e.g., the source decoder, is trained to map the target images to be source-like. We render the target's generation $\widetilde{\bm x}_{ts}$ for the test mode. 
The test image from the target domain $\bm x_t^i$ first passes through the inference model and obtains its deep feature $\bm h_t^i$; then it is fed into the generation model to generate an image $\widetilde{\bm x}_{ts}^i$ with source style while keeping its ground-truth class label $y_t^i$. That is, we try to make the marginal distribution $p(\widetilde{\bm x}_{ts}^i) \approx p(\bm x_s^j)$, and we achieve the classification of $\bm x_t^i$ by classifying $\widetilde{\bm x}_{ts}^i$.

\subsection{Learning in CDLM}
\label{subsec:ch4_learning_in_CDLM}

Our goal is to update the variational parameters to learn a joint distribution and the generation parameters for the desired marginal distribution. Since the latent variables are generated with inputs from both domains, we have a modified formulation adapted from the plain VAEs: 
\begin{equation}
\begin{split}
  \log p(\bm{x}_s, \bm{x}_t) - \text{KL}(q_{\bm \phi}(\bm{\ddot{z}}|\bm{x}_s, \bm{x}_t)\|p(\bm{\ddot{z}}|\bm{x}_s, \bm{x}_t)) \\
  = \mathbb{E}[\log p(\bm{x}_s, \bm{x}_t|\bm{\ddot{z}})] -\text{KL}(q_{\bm \phi}(\bm{\ddot{z}}|\bm{x}_s, \bm{x}_t)\|p(\bm{\ddot{z}})),
\end{split}
\end{equation}
where $\text{KL}(\cdot)$ is the KL divergence, and the transfer latent variable $\bm{\ddot{z}}$ can be either $\bm{\ddot{z}}_{st}$ or $\bm{\ddot{z}}_{ts}$. Minimizing $\text{KL}(q_{\bm \phi}(\bm{\ddot{z}}|\bm{x}_s, \bm{x}_t)\|p(\bm{\ddot{z}}|\bm{x}_s, \bm{x}_t))$ is equivalent to maximizing the variational evidence lower bound (ELBO) $\mathcal{L}(\bm\theta, \bm\phi; \bm x_s, \bm x_t)$:
\begin{equation} \label{eq:ch4_ELBO_kl_model}
\begin{split}
\mathcal{L}(\bm\theta, \bm\phi; \bm x_s, \bm x_t) & = \mathbb{E}_{q_{\bm \phi}}[\log p_{\bm \theta}(\bm{x}_s, \bm{x}_t|\bm{\ddot{z}})] \\
& 
  -\text{KL}(q_{\bm \phi}(\bm{\ddot{z}}|\bm{x}_s, \bm{x}_t)\|p(\bm{\ddot{z}})),
\end{split}
\end{equation}
where the first term corresponds to the reconstruction cost $\mathcal{L}_{\text{Rec}}$, 
and the second term is $\mathcal{L}_{\text{KL}}$, the KL divergence between the learned latent probability and the prior (specified as $\mathcal{N}(0, \bm{I})$). Considering the reconstruction of $\bm{x}_s$, and the KL divergence for both $\bm{\ddot{z}}_{st}$ and $\bm{\ddot{z}}_{ts}$, we have the following loss:
\begin{equation}  \label{eq:ch4_vae_objective}
\begin{split}
\mathcal{L}(\bm\theta, \bm\phi; \bm x_s, \bm x_t) &= \mathbb{E}_{\bm{\ddot{z}} \sim q(\bm{\ddot{z}}_{st}|\bm x_s, \bm x_t)}[\log p_{\bm \theta}(\bm x_s|\bm{\ddot{z}}_{st})] \\
&- \text{KL}(\log q_{\bm\phi}({\bm{\ddot{z}}_{st}|\bm x_s, \bm x_t})\|p(\bm z)) \\
& - \text{KL}(\log q_{\bm\phi}({\bm{\ddot{z}}_{ts}|\bm x_t, \bm x_s})\|p(\bm z)). \\
\end{split}
\end{equation}  

Also, to align the source and target domains' deep representations, we employ an adversarial strategy to regularize the model. The adversarial loss function is given by
\begin{equation}  \label{eq:ch4_adversarial_objective}
  \mathcal{L}_{\text{adv}}= \mathbb{E}_{\bm h_s \sim p(\bm h_s|\bm x_s)}[\log \Xi(\bm h_s)] + \mathbb{E}_{\bm h_t \sim p(\bm h_t|\bm x_t)}[\log (1 - \Xi(\bm h_t))],
\end{equation}
where $\Xi(\cdot)$ is the discriminator to predict from which domain the deep representation feature is generated. 

To improve the inter-class alignment, we also introduce a pairwise consistency regularization between the reconstruction $\mathbf{D}_{\bm \theta}(\bm{\ddot{z}}_{st})$ / $\mathbf{D}_{\bm \theta}(\bm{\ddot{z}}_{ts})$, and the generation for the source $\widetilde{\bm x}_s=\mathbf{D}_{\bm \theta}(\gamma_1\bm h_s + \gamma_2 \bm \epsilon)$ / target $\mathbf{D}_{\bm \theta}(\gamma_1\bm h_t + \gamma_2 \bm \epsilon)$, 
respectively. For the consistency loss $\mathcal{L}_{c}$, both the $L_1$ and $L_2$-norm penalty can be used to regularize the decoder. Here we simply use MSE. Let $\mathcal{L}_c^s$ and $\mathcal{L}_c^t$ be the consistency for the source and target domains respectively. $\mathcal{L}_c$ is given as a combination of these two components, weighted by two coefficients $\beta_1$ and $\beta_2$, as follows:
\begin{equation}  \label{eq:ch4_consistency_objective}
\begin{aligned}
\mathcal{L}_{c} &= \beta_1 \mathcal{L}_c^s + \beta_2 \mathcal{L}_c^t \\
&= \beta_1 [ \mathbb{E}_{\bm{\ddot{z}} \sim q(\bm{\ddot{z}}|\bm{x_s}, \bm{x_t})}\log p(\widehat{\bm{x}}_{s}|\bm{\ddot{z}}_{st})\\
& \ - \mathbb{E}_{\bm{h_s} \sim p(\bm h_s|\bm x_s)} \log p(\widetilde{\bm x}_s|\bm h_s) ] ^2 \\
& \ + \beta_2 [\mathbb{E}_{\bm{\ddot{z}} \sim q(\bm{\ddot{z}}|\bm x_s, \bm x_t)}\log p(\widehat{\bm x}_{ts}|\bm{\ddot{z}}_{ts})\\
& \ - \mathbb{E}_{\bm h_t \sim p(\bm h_t|\bm x_t)} \log p(\widetilde{\bm x}_{ts}|\bm h_t) ]^2. 
\end{aligned}
\end{equation}   

Then, the variational parameters $\bm \phi$ and generation parameters $\bm \theta$ are updated by the following rules:
\begin{equation}  \label{eq:update_weights2}
   \bm \phi \leftarrow \bm \phi - \eta_1 \nabla ( \mathcal{L}_{\text{adv}} + \lambda_1 \mathcal{L}_{\text{KL}} + \lambda_2 \mathcal{L}_{\text{Rec}}),
\end{equation}
and 
\begin{equation}
    \bm \theta \leftarrow \bm \theta - \eta_2 \nabla (\mathcal{L}_{\text{Rec}} + \mathcal{L}_{c}),
\end{equation}
where $\eta_1, \eta_2$ are two learning rates, and $\lambda_1$, $\lambda_2$ are weights used to adjust the impact of the discriminator loss and the reconstruction loss. Note that only data from the desired domain (the source) are used to train the reconstruction loss. The KL term makes the transfer latent space to approximate the prior. We summarize our CDLM in Algorithm ~\ref{algo:ch4_cdlm_algorithm}.

\begin{algorithm}[!thbp]
  \SetAlgoLined 
  \DontPrintSemicolon
  \KwIn{Source: $\mathbf{X}_s$, Target:$\mathbf{X}_t$}
  \KwResult{Inference and Generative model of CDLM}
  $\bm \phi, \bm \theta$ $\leftarrow$ 
  initialization\;
  \For{iterations of traning}{
    $\mathbf{X}_s, \mathbf{X}_t$ $\leftarrow$ sample mini-batch \\
    $\bm \epsilon$ $\leftarrow$ sample from $\mathcal{N}(0, I)$ \\
    $\ddot{\bm z}_{st}, \ddot{\bm z}_{ts}$ $\leftarrow$ sample from $q(\bm{z}_{st}|\mathbf{X}_s, \bm h_t)$ 
    and $q(\pmb{z}_{ts}|\mathbf{X}_t, \bm h_s)$ by Eqs.~(\ref{eq:zst_cdlm_latent}) to (\ref{eq:zts_cdlm_latent}) \\
    Generate ${\widehat{\mathbf{X}}_{st}},{\widehat{\mathbf{X}}_{ts}}, {\widetilde{\mathbf{X}}_s},
    {\widetilde{\mathbf{X}}_t}$ by feeding $\ddot{\bm z}_{st}, \ddot{\bm z}_{ts}$ through Decoders \\ 
    $\mathcal{L}_{\text{Rec}}, \mathcal{L}_\text{KL}, \mathcal{L}_\text{adv}, \mathcal{L}_{c}$ $\leftarrow$ calculated by Eqs.~(\ref{eq:ch4_ELBO_kl_model}) to (\ref{eq:ch4_consistency_objective}) \\
    \For{iterations of inference model updating}{
      $\bm \phi \leftarrow$  -- $\eta_1 \Delta_{\bm \phi} ( \mathcal{L}_{\text{adv}} + \lambda_1 \mathcal{L}_{\text{KL}} + \lambda_2 \mathcal{L}_{\text{Rec}})$\\
    }
    \For{iterations of generative model updating}{
      $\bm \theta$ $\leftarrow$ --$\eta_2 \Delta_{\bm \theta}(\mathcal{L}_{\text{Rec}} + \mathcal{L}_{c})$ \\
    }
  }
  \caption{Training of the CDLM framework}
  \label{algo:ch4_cdlm_algorithm} 
\end{algorithm}

\subsection{Generalization bound analysis}
\label{subsec:ch4_generalization_bound} 

Here we give the generalization error analysis using the domain adaptation theory~\cite{Ben-David2007, Ben-David2010}. The adaptation error on the target,  $\epsilon_t(f)$, is bounded by
\begin{equation}
  \epsilon_t(f) \leq \epsilon_s(f) + \inf \limits_{f^* \in \mathcal{H}}[\epsilon_s(f^*)+\epsilon_t(f^*)] + \hat{d}_{\mathcal{H}}(\mathbf{X}_s, \mathbf{X}_t)
\label{eq:et_upperbound}
\end{equation}
where $f$ is a mapping function of hypothesis family $\mathcal{H}$, $f^*$ is the ideal joint hypothesis to minimize the combined error on source and target predictions, $\epsilon_s(.)$ and $\epsilon_t(.)$ are the prediction error on the source and target data respectively, and  $\hat{d}_{\mathcal{H}}(\cdot)$ is the empirical $\mathcal{H}$-discrepancy between  source and target data. 

In our model, we use the gradient reversal layer (GRL) to align the high-level representations. The objective in Eq.~(\ref{eq:ch4_adversarial_objective})), similar to the treatment in DANN~\cite{Ganin2016}, is equivalent to maximizing
\begin{equation}  \label{eq:ch4_adversarial_objective_transfer_bound}
\begin{aligned}
  \mathcal{L}_{\text{adv+}}(\Xi(\bm h_i^s, \bm h_i^t)) =& -\frac{1}{n}\sum_{i=1}^n d_i^s \log(\Xi(\bm h_i^s)) \\
  & - \frac{1}{n} \sum_{i=1}^n d_i^t \log(1 - \Xi(\bm h_i^t)),
\end{aligned}
\end{equation}
where $d_i^s$ and $d_i^t$ are indicator variables, set to $1$ when the input $\mathbf{x}_i$ is of the source or the target domain respectively. 
This helps reduce the upper bound component $\hat{d}_{\mathcal{H}}(\mathbf{X}_s, \mathbf{X}_t)$ in Eq.~(\ref{eq:et_upperbound}), hence leading to  
a better adaptation performance. 

Furthermore, the other term in Eq.~(\ref{eq:et_upperbound}) for the ideal joint hypothesis $\epsilon_s(f^*)+\epsilon_t(f^*)$ explicitly accounts for adaptability. In our model, we learn a joint latent space $p(\bm \ddot z_{st}, \bm \ddot z_{ts}; \bm x_s, \bm x_t)$ to generate the cross-domain images, that is $p(\mathbf{D}(\bm \ddot z_{st})) \approx p(\mathbf{D}(\bm \ddot z_{ts}))$ with the shared labels. It leads to a smaller $ \epsilon_s(f^*(\mathbf{D}(\bm \ddot z_{st}))) + \epsilon_t(f^*(\mathbf{D}(\bm \ddot z_{ts})))$ term to further reduce the generalization error.

\section{Experiments} \label{sec:experiments}


We have conducted extensive evaluations of CDLM in two homogeneous transfer scenarios including unsupervised domain adaptation and image-to-image translation. Our model is implemented using TensorFlow~\cite{Abadi2016}. The structures of the encoder and the decoder adopt those of UNIT~\cite{Liu2017} which perform well for image translation tasks. A two-layer fully connected MLP is used for the discriminator. SGD with momentum is used for updating the variational parameters, and Adam for updating generation parameters. The batch size is set to 64. During the experiments, we set  $\gamma_1=1.0,\gamma_2=0.1$,   $\lambda_1=\lambda_2=0.0001$, $\beta_1=0.1$ and $\beta_2=0.01$. 

\subsection {Datasets}
\label{subsec:ch4_datasets}
We have evaluated our model on a variety of benchmark datasets including
MNIST~\cite{LeCun1998}, MNISTM~\cite{Ganin2016}, USPS~\cite{LeCun1989}, Fashion-MNIST~\cite{Xiao2017}, LineMod~\cite{Hinterstoisser2012,Wohlhart2015}, Zap50K-shoes~\cite{Yu2017} and CelebA~\cite{Liu2015,Liu2018}.

\begin{itemize}
\item \textbf{MNIST:} The MNIST handwritten dataset~\cite{LeCun1998} is a very popular machine learning dataset. It has a training set of 60,000 binary images and a test set of 10,000. There are 10 classes in the dataset. In our experiments, we use the standard split of the dataset. MNISTM~\cite{Ganin2016} is a modified version for the MNIST, with random RGB background cropped from the Berkeley Segmentation Dataset\footnote{URL \url{https://www2.eecs.berkeley.edu/Research/Projects/CS/vision/bsds/}}.

\item 
\textbf{USPS:} USPS is a handwritten zip digits datasets~\cite{LeCun1989}.  It contains 9298 binary images ($16\times16$), 7291 used as the training set, while the remaining 2007 are used as the test set. The USPS samples are resized to $28\times28$, the same as MNIST.  

\item 
\textbf{Fashion:} Fashion~\cite{Xiao2017} contains 60,000 images for training, and 10,000 for testing. All the images are greyscaled, $28\times28$ in size space. Besides, following the protocol in~\cite{Ganin2016}, we add random noise to the Fashion images to generate the FashionM dataset, with random RGB background cropped from the Berkeley Segmentation Dataset. 

\item \textbf{LineMod 3D images} For this scenario, 
followed the protocol of~\cite{Bousmalis2017}, we render the LineMod~\cite{Hinterstoisser2012,Wohlhart2015} for the adaptation between synthetic 3D images (source) and real images (target). We note them as L-3D and L-Real in short. The objects with different poses are located at the centre of the images. The synthetic 3D images render a black background and a variety of complex indoor environments for real images. We use the RGB images only, not the depth images. 
 
\item 
\textbf{CelebA:} CelebA~\cite{Liu2015} is a large celebrities face image dataset. It contains more than 200K images annotated with 40 facial attributes. We select 50K images randomly, then transform them to sketch images followed the protocol of~\cite{Liu2018}. The original and sketched images are used for translation. 

\item 
\textbf{UT-Zap50K-shoes:} This dataset~\cite{Yu2017} contains 50K shoes images with 4 different classes. During the translation, we get the edges produced by the Canny detector~\cite{Canny1986}. 

\end{itemize}


\subsection{Quantitative results on unsupervised domain adaptation}
\label{subsec:ch4_quantitative_results_UDA}

The CDLM model is applied to unsupervised domain adaptation tasks, where a classifier is also trained using labelled samples in the source domain. 
We choose DANN~\cite{Ganin2016} as the baseline, and compare our model with several other state-of-the-art domain adaptation methods, including
Conditional Domain Adaptation Network (CDAN)~\cite{Long2018}, Pixel-level Domain Adaptation (PixelDA)~\cite{Bousmalis2017}, Unsupervised Image-to-Image translation (UNIT)~\cite{Liu2017}, Cycle-Consistent Adversarial Domain Adaptation (CyCADA)~\cite{Hoffman2018}, Generate to Adapt (GtA)~\cite{Sankaranarayanan2018}, Transferable Prototypical Networks (TPN)~\cite{Pan2019}, Domain Symmetric Networks (SymmNets-V2)~\cite{Zhang2020a}, Instance Level Affinity-based Networks (ILA-DA)~\cite{Sharma2021}, and Deep Adversarial Transition Learning (DATL)~\cite{Hou2022}. 
We also use the source-only and target-only training outcome as the lower and upper bounds following the practice in~\cite{Bousmalis2017, Ganin2016}. 
 
CDLM can transfer the target images to be source-like. Then, for the adaptation tasks, joint or decoupled task learning can be used, and a supervised classifier trained on the source images along with their labels can also be applied to recognize the target images. We adopt a joint task classifier. The parameters of the task classifier $\bm \theta_T$ are jointly learned with the update of $\bm \phi$ and $\bm \theta$. The performance of domain adaptation for the different tasks is shown in Table~\ref{table:ch4_accuracy}. There are 4 scenarios and 7 tasks. Each scenario has bidirectional tasks for adaptation except LineMod. For LineMod, it is adapted from a synthetic 3D image to real objects. For the same adaptation task, we cite the accuracy from the corresponding references; otherwise, some tasks' accuracy is obtained by training the open-source code provided by authors with suggested optimal parameters for a fair comparison. 

From Table~\ref{table:ch4_accuracy}, we can see that our method has an obvious advantage over the baseline and the source-only (lower bound) accuracy, and its performance is only a little lower than the target-only accuracy (upper bound) on both adaptation directions. In comparison with other models, our model has the best performance for most tasks. The CDLM has higher adaptation accuracies for the scenarios with a seemingly large domain gaps, such as MNIST$\rightarrow$MNISTM and Fashion$\rightarrow$FashionM. For the 3D scenario, our model's performance is a little lower than PixelDA~\cite{Bousmalis2017} and DATL~\cite{Hou2022} but outperforms all the other methods. In PixelDA, the input is not only source images but also depth image pairs. It might be helpful for the generation. Also, the results show that DATL has better performance for some tasks such as MNISTM$\rightarrow$MNIST and LineMod3D to Real. This is possibly due to the better intermediate generation by DATL in these scenarios, but its computational complexity is much higher.

\begin{table*}[!thbp]
  \centering
  \caption[\small Mean classification accuracy comparison.]{\small Mean classification accuracy comparison. The ``source only" row is the accuracy for target without domain adaptation training only on the source. The ``target only" is the accuracy of the full adaptation training on the target. For each source-target task the best performance is in bold} 
  \resizebox{0.95\textwidth}{!}{
    \begin{tabular}{|l|c|c|c|c|c|c|c|c|c|}
      \hline
      Source & MNIST & USPS & MNIST& MNISTM & Fashion & FashionM & LineMod 3D\\
      Target & USPS  & MNIST &MNISTM & MNIST& FashionM & Fashion & LineMod Real\\
      \hline\hline
      Source Only & 0.634 & 0.625 & 0.561  & 0.633  & 0.527 & 0.612 & 0.632 \\
      \hline\hline
      DATL~\cite{Hou2022} & \textbf{0.961} & 0.956 & 0.890 & \textbf{0.983} & 0.853 & 0.917 & \textbf{0.998} \\ 
      SymmNets-V2~\cite{Zhang2020a} & 0.948 & 0.968 & - & - & - & - & - \\ 
      ILA-DA~\cite{Sharma2021} & 0.949 & 0.975 & - & - & - & - & - \\ 
      DANN~\cite{Ganin2016} & 0.774 & 0.833 & 0.766  &  0.851   &  0.765  & 0.822 & 0.832  \\
      CyCADA~\cite{Hoffman2018} & 0.956  & 0.965 &  0.921  &  0.943  & 0.874 & 0.915 &0.960 \\
      GtA~\cite{Sankaranarayanan2018} & 0.953  & 0.908  &  0.917 &  0.932 & 0.855 & 0.893 & 0.930  \\
      CDAN~\cite{Long2018} & 0.956 & 0.980 & 0.862 & 0.902 & 0.875 & 0.891 & 0.936  \\
      PixelDA~\cite{Bousmalis2017} & 0.959 & 0.942 &  0.982 &  0.922  & 0.805   &  0.762 & \textbf{0.998} \\
      TPN~\cite{Pan2019} & 0.921 & 0.941 & - & - & - & - & - \\ 
      UNIT~\cite{Liu2017} & 0.960 & 0.951  & 0.920  & 0.932 & 0.796   &  0.805 & 0.964 \\ 
      
      \hline\hline
      CDLM ($\widetilde{\bm x}_{ts}$) & \textbf{0.961}  & \textbf{0.983} & \textbf{0.987}   & 0.962  &   \textbf{0.913} & \textbf{0.922} & 0.984  \\
      \hline \hline
      Target Only & 0.980 &  0.985 & 0.983  & 0.985  &  0.920  &  0.942 & 0.998   \\
      \hline  
    \end{tabular}}
  \label{table:ch4_accuracy}
\end{table*}

\subsection {Qualitative results on unsupervised domain adaptation }
\label{subsec:ch4_qualitative_results_UDA}

\begin{figure}[!thbp]
  \centering
  \begin{subfigure}[t]{.48\textwidth}
    \centering
    \includegraphics[width=0.90\textwidth]{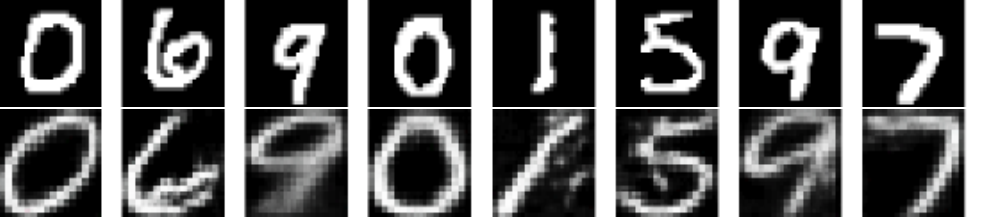} %
    \caption{\small USPS$\rightarrow$MNIST}
    \label{fig:ch4_usps_mnsit}
  \end{subfigure}
  \begin{subfigure}[t]{.48\textwidth}
    \centering
    \includegraphics[width=0.90\textwidth]{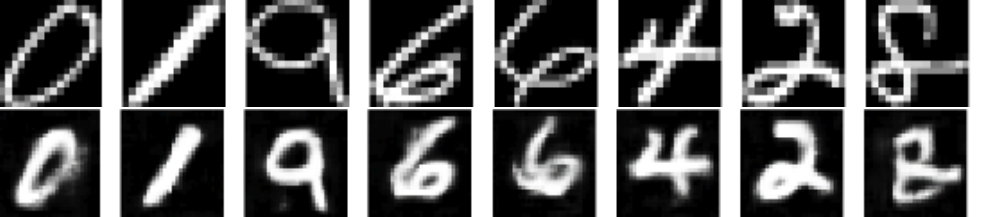} %
    \caption{\small MNIST$\rightarrow$USPS}
    \label{fig:ch4_mnist_usps}
  \end{subfigure}
  \begin{subfigure}[t]{.48\textwidth}
    \centering
    \includegraphics[width=0.90\textwidth]{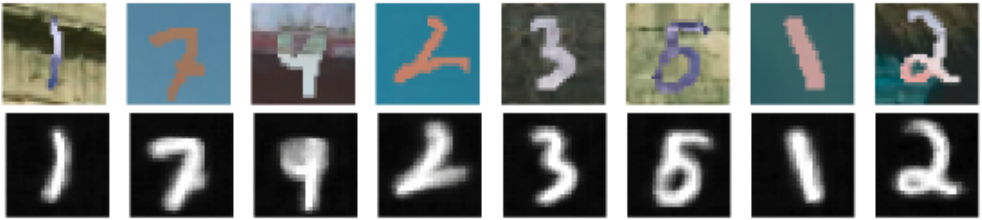}
    \caption{\small MNIST$\rightarrow$MNISTM}
    \label{fig:ch4_mnist_mnistm}
  \end{subfigure}
  \begin{subfigure}[t]{.48\textwidth}
    \centering
    \includegraphics[width=0.90\textwidth]{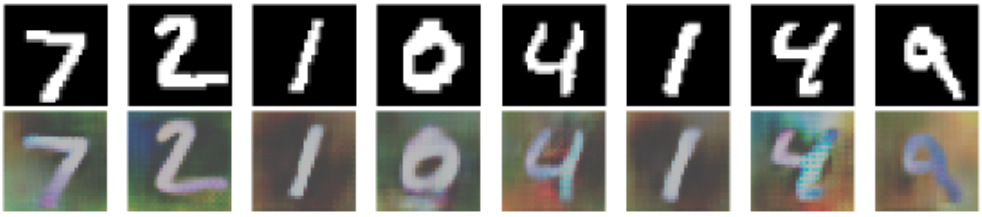}
    \caption{\small MNISTM$\rightarrow$MNIST}
    \label{fig:ch4_mnistm_mnist}
  \end{subfigure}
  \begin{subfigure}[t]{.48\textwidth}
    \centering
    \includegraphics[width=0.90\textwidth]{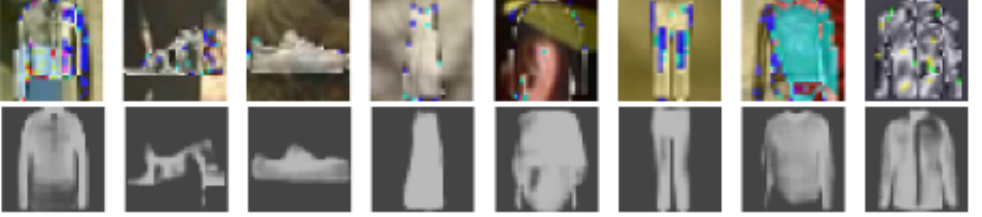}
    \caption{\small Fashion$\rightarrow$FashionM}
    \label{fig:ch4_fashion_fashionm}
  \end{subfigure}
  \begin{subfigure}[t]{.48\textwidth}
    \centering
    \includegraphics[width=0.90\textwidth]{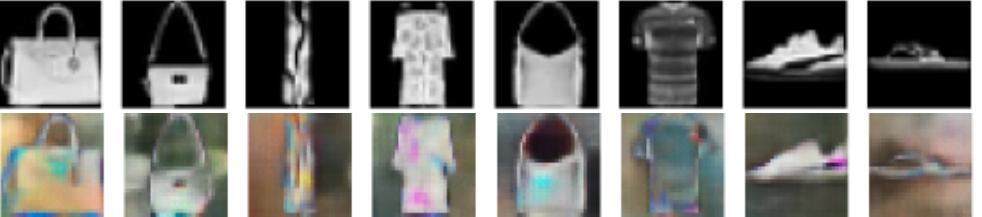}
    \caption{\small FashionM$\rightarrow$Fashion}
    \label{fig:ch4_fashionm_fashion}
  \end{subfigure}
  \caption[\small Visualization for the adaptations.]{\small Visualization for the adaptations. 6 different tasks are illustrated. For each task, the first row shows target images and the second row shows the adapted images with source-like style.}
  \label{fig:ch4_digits_and_fashion}   
\end{figure}

We now present the visual assessment of the domain adaptation outcomes, showing the original target images and their CDML generated counterparts using the \textit{style} of the source domain. Fig.~\ref{fig:ch4_digits_and_fashion} is the visualization for the digits and Fashion adaptation, respectively. 
The generation for the task USPS$\rightarrow$MNIST is shown in Fig.~\ref{fig:ch4_usps_mnsit}, where we can see target MNIST images are transferred to the USPS style while its semantic content. For example, the digit ``1'' in MNIST becomes more leaned than its original look, and ``9'' becomes flattened. In Fig.~\ref{fig:ch4_mnist_usps}, the target USPS digits gain the MNIST style. For scenarios in Fig.~\ref{fig:ch4_mnist_mnistm} and Fig.~\ref{fig:ch4_mnistm_mnist}, colour backgrounds are effectively removed or added through adaptation, respectively.

For the Fashion scenario, fashion items have a variety of different textures and content. Besides, the noisy backgrounds pollute the items randomly. For example, different parts of cloth may have different colours. 
Fig.~\ref{fig:ch4_fashion_fashionm} is for the task Fashion $\rightarrow$ FashionM,  where the noisy background is removed while the content is kept. On the other hand, Fig.~\ref{fig:ch4_fashionm_fashion} shows that the original target Fashion images are added with a similar noisy background as the source. All these indicate very promising adaptation outcome. 

\begin{figure*}  
  \centering
  \includegraphics[width=0.8\textwidth]{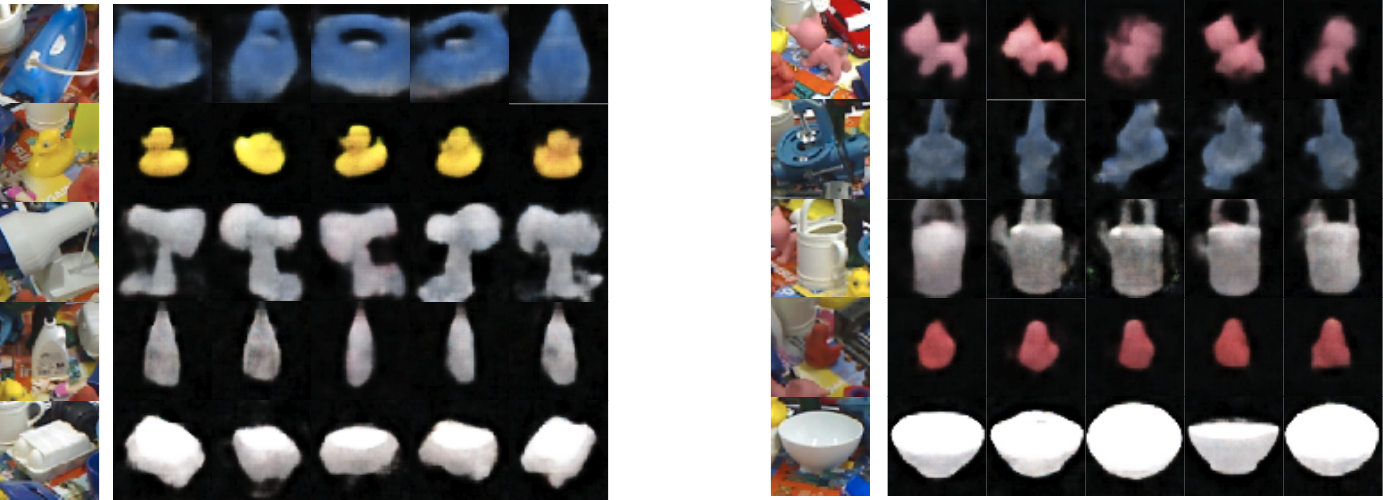}
  \caption[\small Visualization of task LineMod 3D Synthetic$\rightarrow$Real.]{\small Visualization of task LineMod 3D Synthetic$\rightarrow$Real. For a query image (on the left), different adaptation images (to the right) with various poses can be generated.}
  \label{fig:ch4_linemod}
\end{figure*}

For the scenario of LineMod3D, the images of the real objects with different backgrounds are transferred to the source (synthetic 3D) images with a black background. We only use the image's RGB channel, and the depth image is not included in the inputs during the evaluation. Our CDLM can generate realistic synthetic images for the adaptation. Due to the 3D style, the generation of the target gives different poses. For example, in Fig.\ref{fig:ch4_linemod}, different poses of the iron object are obtained from different trials.

\begin{table*}[!h]
  \centering
  \caption[\small Evaluation on the effect of unsupervised consistency metrics.]{\small Evaluation on the effect of unsupervised consistency metrics. The recognition accuracy is shown for four tasks in the unsupervised domain adaptation scenario. Our model is on the last row with both the $\mathcal{L}_c^s$ and $\mathcal{L}_c^t$, which achieves the best performance.}
    \begin{tabular}{|l|c|c|c|c|}
      \hline
      Model/Tasks & MNIST$\rightarrow$USPS & USPS$\rightarrow$MNIST & Fashion$\rightarrow$FashionM & FashionM$\rightarrow$Fashion \\
      \hline \hline
      CDLM w/o $\mathcal{L}_c$ & 0.635 & 0.683 & 0.646 & 0.672  \\
      CDLM+$\mathcal{L}_c^t$ & 0.689 & 0.695 & 0.682 & 0.691  \\
      CDLM+$\mathcal{L}_c^s$ & 0.951 & 0.980 & 0.912 & 0.915  \\
      CDLM+$\mathcal{L}_c^s$+$\mathcal{L}_c^t$  & 0.961 & 0.983 & 0.913 & 0.922  \\
      \hline   
    \end{tabular}
  \label{table:ch4_wo_consistency_loss}
\end{table*}

\begin{figure}
  \begin{subfigure}[t]{0.4\textwidth}
    \centering
    \includegraphics[width=0.6\textwidth]{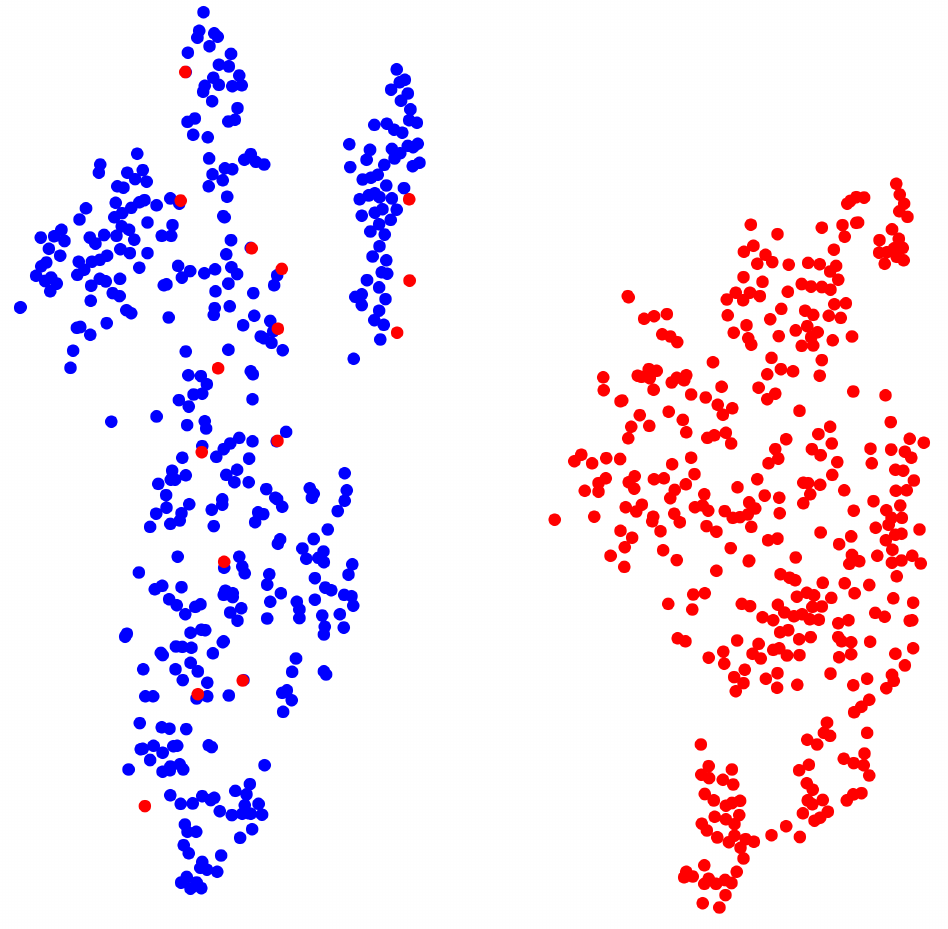}
    \caption{\small Visualization of $\bm x_s (blue), \bm x_t$ (red), MNISTM $\rightarrow$ MNIST.}
    \label{fig:ch4_tsne_mnistm_mnist_h}
  \end{subfigure}
  \hfill
  \begin{subfigure}[t]{0.4\textwidth}
    \centering
    \includegraphics[width=0.6\textwidth]{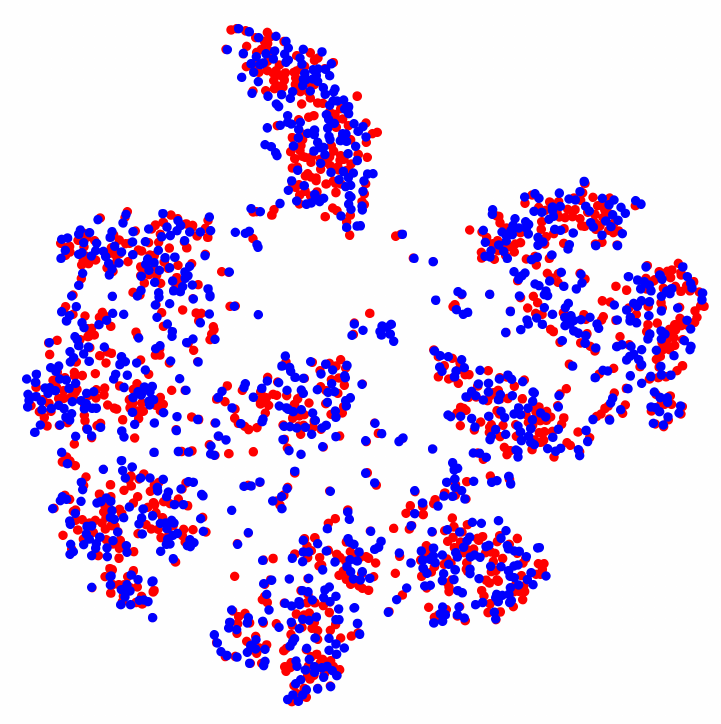}
    \caption{\small Visualization of $\bm{\ddot{z}}_{st}, \bm{\ddot{z}}_{ts}$,  MNISTM $\rightarrow$ MNIST: $\bm{\ddot{z}}_{st}$ in blue, $\bm{\ddot{z}}_{ts}$ in red.}
    \label{fig:ch4_tsne_mnistm_mnist_z}
  \end{subfigure}
  \begin{subfigure}[t]{0.4\textwidth}
    \centering
    \includegraphics[width=0.6\textwidth]{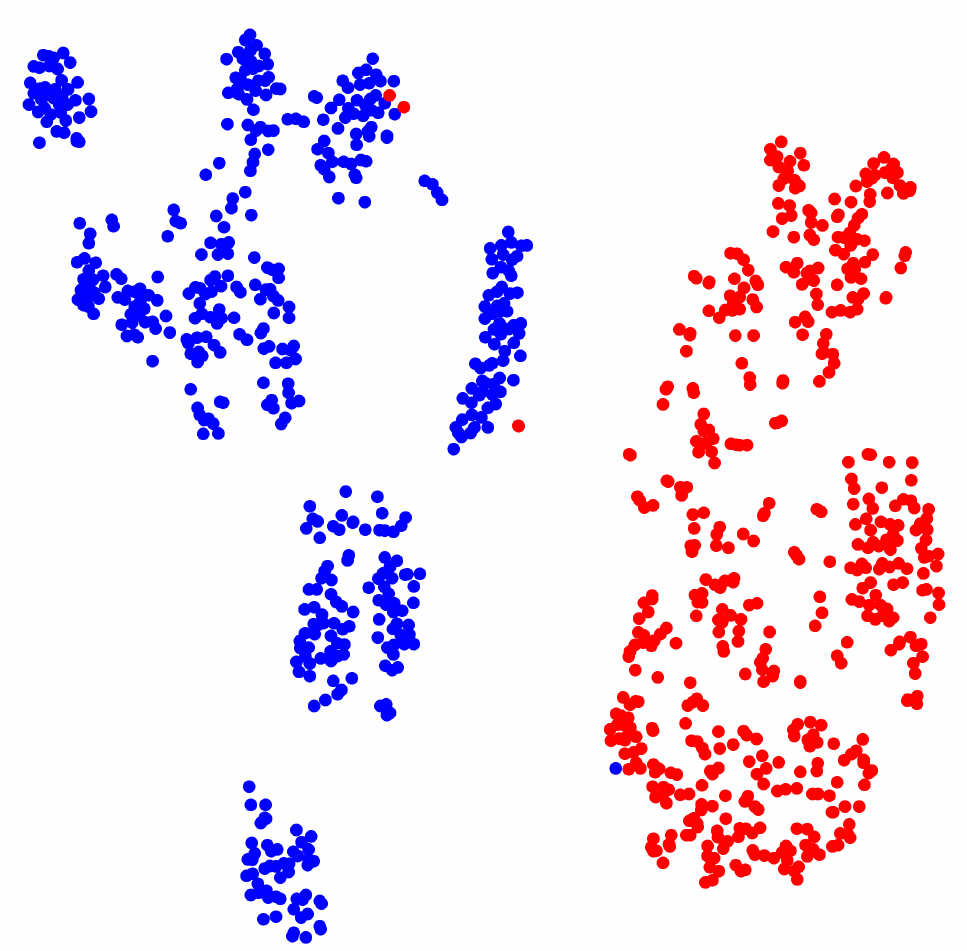}
    \caption{\small Visualization of $\bm x_s (blue), \bm x_t$ (red), MNIST $\rightarrow$ USPS.}
    \label{fig:ch4_tsne_mnist_usps_h}
  \end{subfigure}
  \hfill
  \begin{subfigure}[t]{0.4\textwidth}
    \centering
    \includegraphics[width=0.6\textwidth]{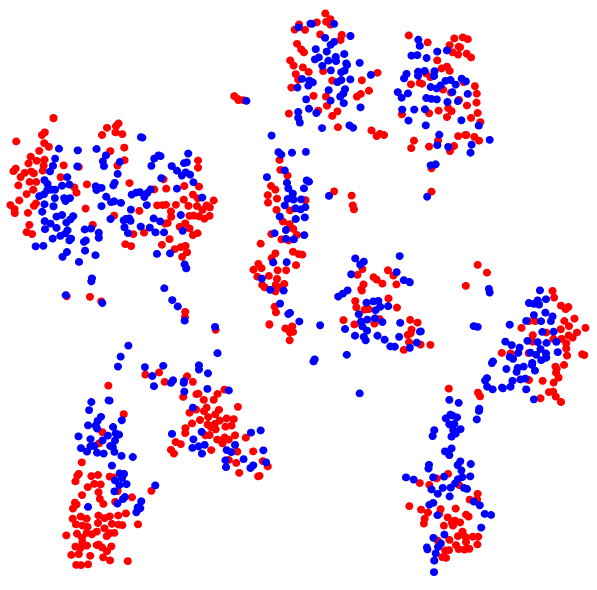}
    \caption{\small Visualization of $\bm{\ddot{z}}_{st}, \bm{\ddot{z}}_{ts}$,  MNIST $\rightarrow$ USPS: $\bm{\ddot{z}}_{st}$ in blue, $\bm{\ddot{z}}_{ts}$ in red.}
    \label{fig:ch4_tsne_mnist_usps_z}
  \end{subfigure}
  \caption{\small t-SNE visualization of deep representations and modulated latent encodings. Both embeddings align well between the source and the target. The deep representations is demonstrated with data distribution information to support the modulations.}
  \label{fig:ch4_tSNE}
\end{figure}

Next, we investigate the visualization of the latent embedding of the high-dimensional representations using the t-SNE visualization~\cite{Maaten2008} of deep representations ($\bm h_s, \bm h_t$) and latent encodings ($\bm{\ddot{z}}_{st}, \bm{\ddot{z}}_{ts}$) w.r.t source and target, respectively. Fig.~\ref{fig:ch4_tsne_mnistm_mnist_h} - \ref{fig:ch4_tsne_mnistm_mnist_z} are the visualizations for the task MNISTM-MNIST and \ref{fig:ch4_tsne_mnist_usps_h} - \ref{fig:ch4_tsne_mnist_usps_z} for the task MNIST-USPS. 
Both visualizations demonstrate a good alignment between the source and target data resulted in the cross-modulated latent space, 
confirming the effectiveness of our algorithm design.

\subsection{Results on cross-domain image translation}
\label{subsec:ch4_UIT}

\begin{figure}[!thbp]
  \centering
  \begin{subfigure}{.8\columnwidth}
    \centering
    \includegraphics[width=1.0\textwidth]{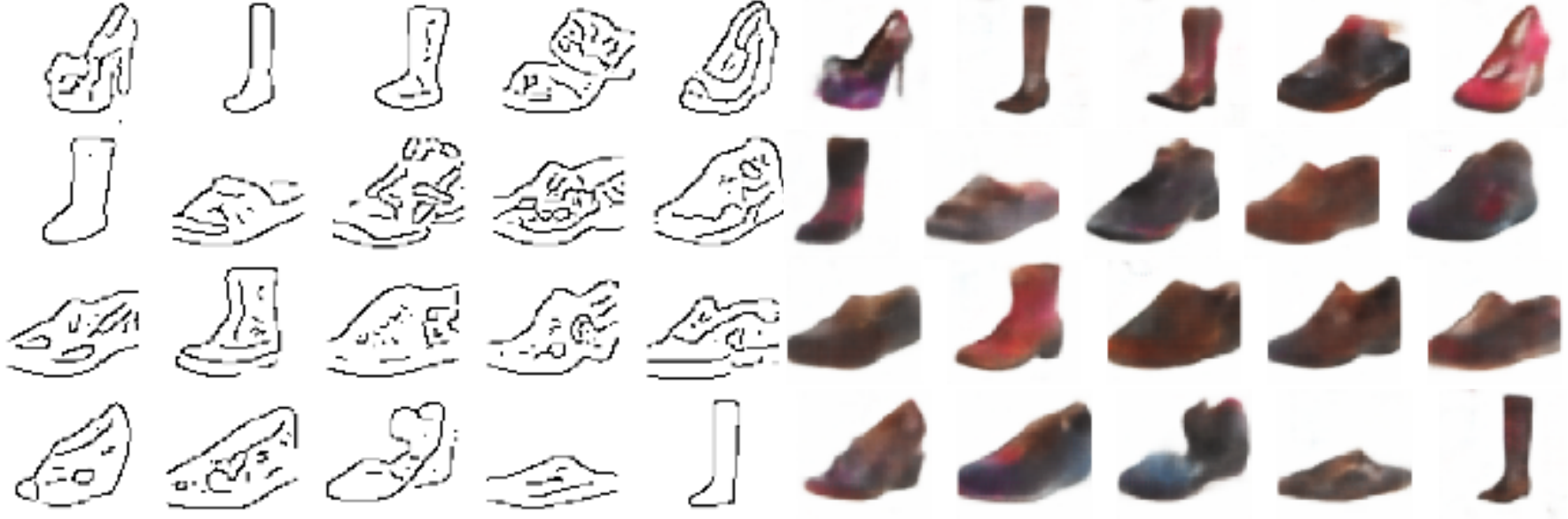} %
    \caption{\small ``edge'' to ``shoes''}
    \label{fig:ch4_edge_shoes}
  \end{subfigure} \hfill
  \begin{subfigure}{.8\columnwidth}
    \centering
    \includegraphics[width=1.0\textwidth]{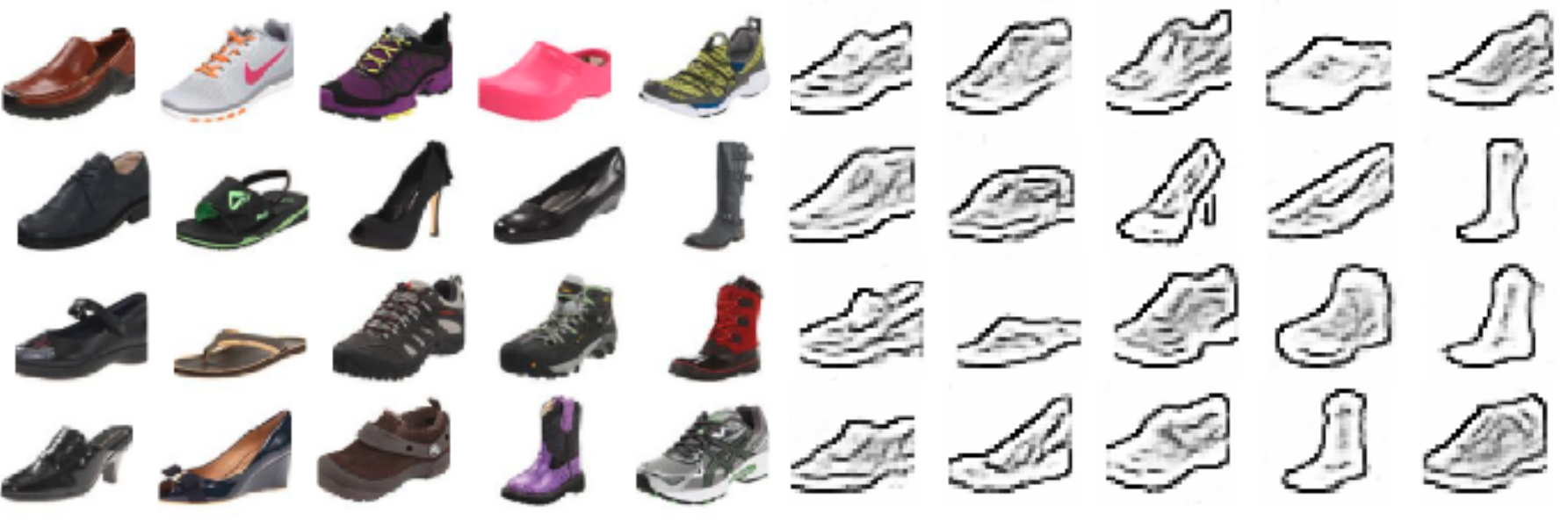} %
    \caption{\small ``shoes'' to ``edge''}
    \label{fig:ch4_shoes_edge}
  \end{subfigure} \hfill
  \begin{subfigure}{.8\columnwidth}
    \centering
    \includegraphics[width=1.0\textwidth]{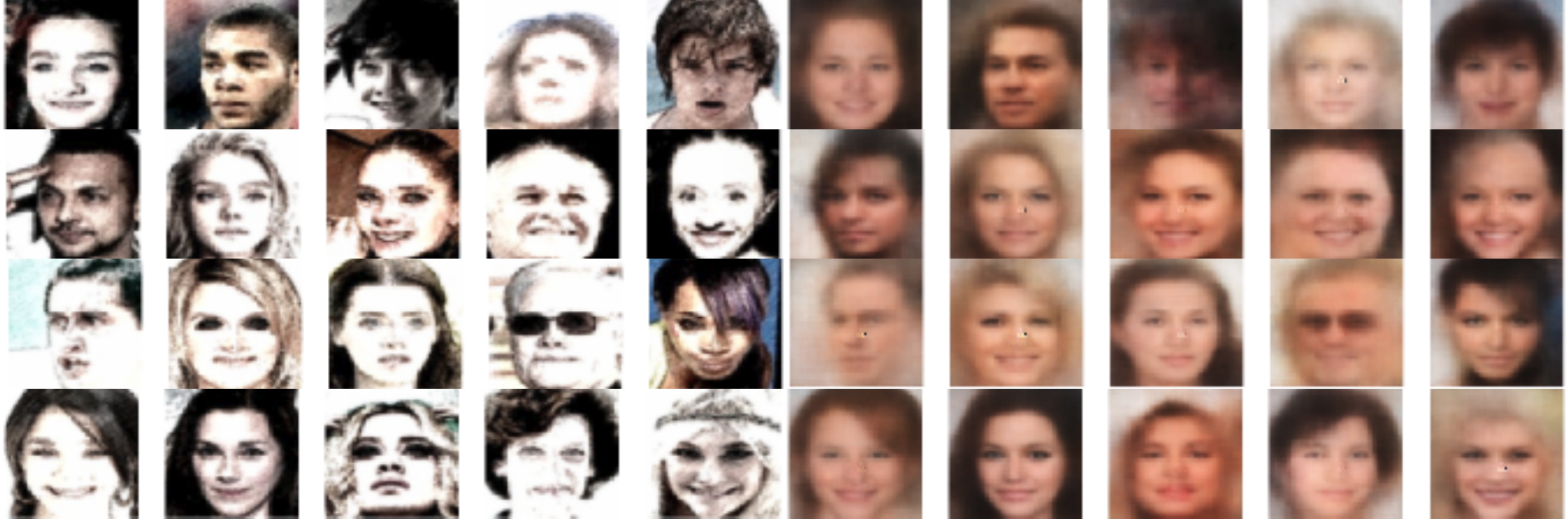}  %
    \caption{\small ``sketch'' to ``face''}
    \label{fig:ch4_sketch_face}
  \end{subfigure} \hfill
  \begin{subfigure}{.8\columnwidth}
    \centering
    \includegraphics[width=1.0\textwidth]{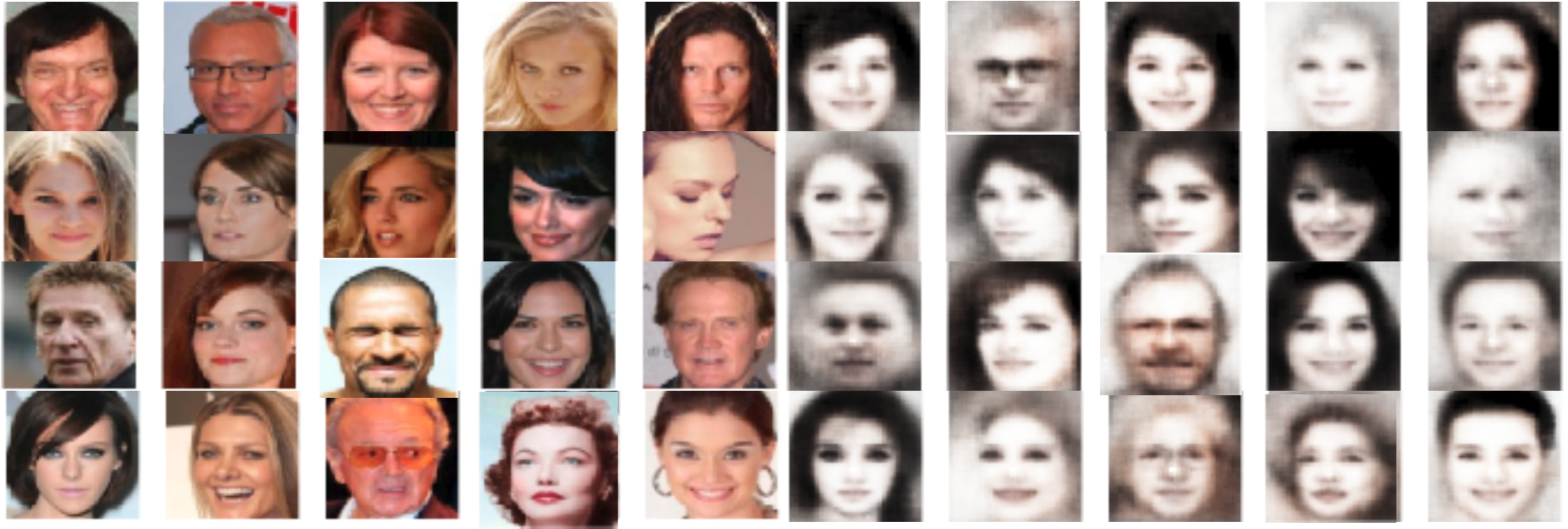} %
    \caption{\small ``face'' to ``sketch''}
    \label{fig:ch4_face_sketch}
  \end{subfigure}
  \caption{\small Visualization of cross-domain image mapping. }
  \label{fig:ch4_image_trans}   
\end{figure}

The proposed model also can be used for cross-domain image translation. Fig.~\ref{fig:ch4_image_trans} gives the demonstration for the mapping experiments. Specifically, Fig.~\ref{fig:ch4_edge_shoes} and~\ref{fig:ch4_shoes_edge} are for shoes and edges; Fig.~\ref{fig:ch4_sketch_face} and~\ref{fig:ch4_face_sketch} are for faces and pencil sketches. We can see that the proposed model can effectively translate the edges to their counterpart  for shoes and edges. The translation is multi-modal, which means the edges can generate different colour shoes with different tests. For the challenging face and pencil sketch task, the proposed model can also map one to another on both directions. The generations have some variations compared with the original images, but in general our method can generate realistic, translated images. 

Compared with the ``sketch'' to ``real'' conversion, the reverse task seems trickier. For example, when the face image is given, the generated sketch may lose some details. The reason may be that the low-level features are neglected when the deep feature acts as the condition. Another observation is the blurriness in images. It is caused by the intrinsic character of VAEs that employs a $L_2$-based regularizer for the marginal evidence. 

Furthermore, we employ a few quantitative performance metrics for image translation evaluation. SSIM~\cite{Wang2004}, MSE, and PSNR are used for the evaluation. The results are shown in Table~\ref{table:ch4_image_mapping_quantitative_results}. We can see that our model outperforms E-CDRD~\cite{Liu2018}, which learns a disentangled latent encoding for the source and the target for domain adaptation. Meanwhile, it matches the performance of StarGAN~\cite{Choi2018}, which is designed for multi-domain image translation. 

\begin{table}[!thbp] 
  \centering
  \caption{\small Performance for the image mapping.}
  \begin{tabular}{|c|c|c|c|}
    \hline
    \multirow{2}{*}{Models} & \multicolumn{3}{c|}{``Sketch'' to ``Face''} \\
    \cline{2-4}
    & SSIM  &  MSE & PSNR  \\
    \hline \hline
    E-CDRD~\cite{Liu2018} & 0.6229  & 0.0207 & 16.86 \\
    StarGAN~\cite{Choi2018} & \textbf{0.8026} & 0.0142 &  19.04 \\
    \hline
    CDLM & 0.7961 & \textbf{0.0140} &  \textbf{19.89}  \\
    \hline
  \end{tabular}
  \label{table:ch4_image_mapping_quantitative_results}
\end{table}

In addition, some classification experiments are carried out to evaluate the translation performance. We take ``shoes'' as an example, which are labelled into four different classes. Our proposed model's recognition accuracy for task shoes$\rightarrow$edge is 0.953, which is higher than the results of PixelDA (0.921) and UNIT (0.916), respectively.

\subsection{Ablation studies}
\label{subsec:ch4_ablation_study}

In our model, two unsupervised consistency metrics are added for achieving good generation quality. Ablation studies are then carried out, where adaptation accuracy is used for evaluation. Table~\ref{table:ch4_wo_consistency_loss} gives the results for four different tasks. The performance without $\mathcal{L}_c$ drops down because the decoder cannot generate realistic cross-domain images. $\mathcal{L}_c^t$ connects outputs generated from $\bm h_t$ and $\ddot{\bm z}_{ts}$ only for the target, which improves the performance slightly. Meanwhile, we can see that the $\mathcal{L}_c^s$ loss boosts the accuracy for adaptation significantly, which connects the two domains with the generations by $\bm h$. Finally, the scenario with both $\mathcal{L}_c^s$ and $\mathcal{L}_c^t$ gives the best performance in all four tasks. It bridges both the feature representations $\bm h$ and the latent representations $\ddot{\bm z}$ between the two domains.

\subsection {Model analysis}
\label{subsec:ch4_model_analysis}
We further examine the effect of varying settings on CDML learning outcome. 
\noindent
\textbf {The effect of the encoder layers:} In the CDML model, deep features play an instrumental role and are utilized to cross-modulate the transfer latent encoding. They  may be potentially influenced by the encoder's depth. To look into this, we use MNIST $\rightarrow$ USPS and Fashion $\rightarrow$ FashionM for evaluation. For the first task, they have different contents, but with the same background. The second task is  different, images have the same content but different backgrounds. The outputs of different encoder layers (Conv4, Conv5 and Conv$_{last}$) are used for the experiments. 

\begin{table} [!hbtp]
  \centering
  \caption[\small Adaptation accuracy with different layer depth.]{\small Adaptation accuracy with different layer depth. Tasks MNIST $\rightarrow$ USPS and Fashion $\rightarrow$ FashionM are considered.}
    \begin{tabular}{|c|c|c|c|}
      \hline
      \small Tasks/Layers & \small Conv4  & \small Conv5  &  \small Conv\textsubscript{last}  \\
      \hline
      \small MNIST $\rightarrow$ USPS & 0.954  & 0.956  & 0.961   \\
      \small Fashion $\rightarrow$ FashionM& 0.890 & 0.905  &  0.913  \\
      \hline
    \end{tabular}
  \label{table:ch4_different_layers_performance} 
\end{table}

\begin{table*}[!thbp]
  \centering
  \caption[\small Adaptation accuracy with different ($\gamma_1, \gamma_2$).]{\small Adaptation accuracy with different ($\gamma_1, \gamma_2$). Tasks MNIST $\rightarrow$ USPS and Fashion $\rightarrow$ FashionM are considered.}
    \begin{tabular}{|c|c|c|c|c|c|}
      \hline
      \small Tasks / ($\gamma_1,\gamma_2$) & (\small 0.1,1.0)  &  (\small 0.5,0.5)  &  (\small 0.9,0.1) & (\small 1.0,0.1) & (\small 1.0, 0) \\
      \hline \hline
      \small MNIST $\rightarrow$ USPS & {0.320}  & {0.723}  & 0.961 & 0.961 & 0.961 \\
      \small Fashion $\rightarrow$ FashionM & {0.226} & {0.513}  &  0.912 & 0.913 & 0.913 \\
      \hline
    \end{tabular}
  \label{table:ch4_different_gamma}
\end{table*}

The results shown in  Table~\ref{table:ch4_different_layers_performance} indicate that higher accuracies are achieved when more layers are used to extract the deep representations. 
This is expected, since features extracted by higher layers would naturally eliminate lower-level variations between domains, such as background and illumination changes in the images. The performance achieved by representations from different layer depths, however, is quite close. 

\noindent
\textbf{Effect of cross-modulation weighting}. To experiment with $\gamma_1, \gamma_2$, we use the last convolutional layer for the deep representations and evaluate the different weighting values. From Table~\ref{table:ch4_different_gamma}, we can see that the performance drops down significantly with a smaller $\gamma_1$ compared with $\gamma_2$, and increased with a larger $\gamma_1$. The performance seems to be stabilized when $\gamma_1$ is greater than 0.9, while $\gamma_2$ remains 0.1. Following the standard VAE, we keep the noise $\bm \epsilon$ ($\gamma_2\neq 0$) in evaluation. Meanwhile, our model works well even when $\gamma_2=0$. The results suggest that deep representation plays a crucial role in cross-domain modulation. It is also consistent with what we have found in Sec.~\ref{sec:KLDana}: $\gamma_1>0.5$ is preferred and generating more competitive performance.

\noindent
\textbf{Performance of partial modulation:} The proposed transfer latent space is a generalization of the transfer space. Several transfer functions can be satisfied for this purpose. We give the performance of the two partial modulation variants described in Section~\ref{subsec:ch4_cdlm_framework}, as shown in Table~\ref{table:ch4_incomplete_modulation_variants_performance}. The results show that the CDLM, with full latent modulations has better performance than the two partial modulation variants. This is possibly because the partial modulation causes that only the second-order moments contribute to the transfer and cross-consistency, limiting the development of adaptability.

\begin{table}[!htbp]
  \centering
  \caption{\small Performance of the partial modulation variants.}
    \begin{tabular}{|c|c|c|}
      \hline
      Model & MNIST$\rightarrow$MNISTM & MNIST$\rightarrow$USPS \\
      \hline \hline
      CDLM-PM1 & 0.959 & 0.950    \\
      CDLM-PM2 & 0.967 &  0.953   \\
      \hline
      CDLM & 0.987  & 0.961   \\
      \hline
    \end{tabular}
  \label{table:ch4_incomplete_modulation_variants_performance}
\end{table} 

\begin{figure}[!htbp]
  \centering
  \includegraphics[width=0.7\columnwidth]{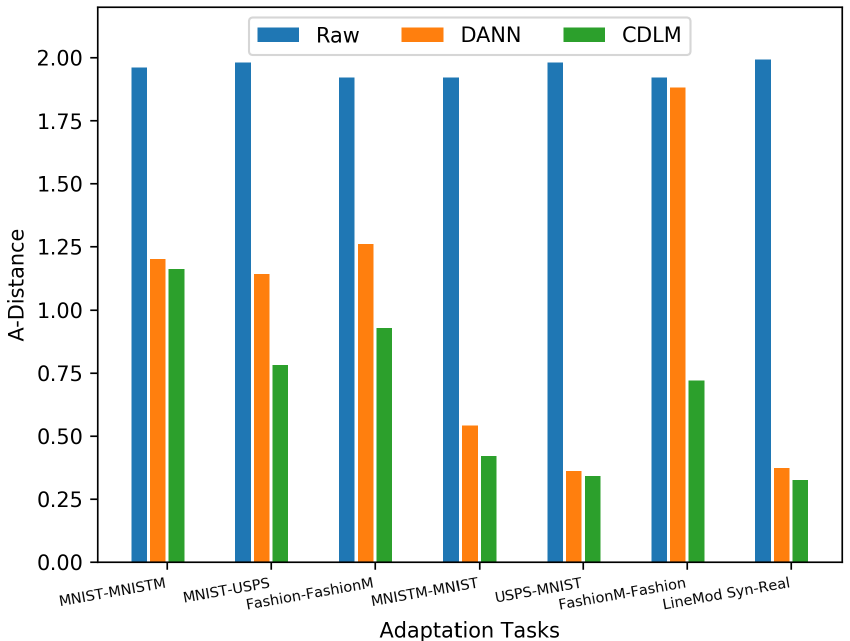}
  \caption{\small $\mathcal{A}$-distances comparison for four tasks.}
  \label{fig:ch4_A_distance}
\end{figure}

\noindent
\textbf{$\mathcal{A}$-Distance:} 
$\mathcal{A}$-distance can be used as a measure for domain discrepancy~\cite{Ben-David2007}. As the exact $\mathcal{A}$-distance is intractable, a proxy is defined as $\hat{d}_{\mathcal{A}} = 2(1 - 2\epsilon)$, where $\epsilon$ is the generalization error of a binary classifier (e.g., kernel SVM) trained to distinguish the input's domain (source or target). Following the protocol of~\cite{Long2015, Peng2020}, we calculate the $\mathcal{A}$-distance on four adaptation tasks under the scenarios of the original images (``Raw''), and the adapted features in DANN and CDLM. The results are shown in Fig.~\ref{fig:ch4_A_distance}. We observe that both DANN and CDLM reduce the domain discrepancy compared with the Raw images scenario, and the $\mathcal{A}$-distance of CDLM is smaller than DANN's. This suggests the domain discrepancy between the source and the target generations gets most reduced by CDLM. 

\begin{figure}[!htbp]
  \centering
  \includegraphics[width=0.7\columnwidth]{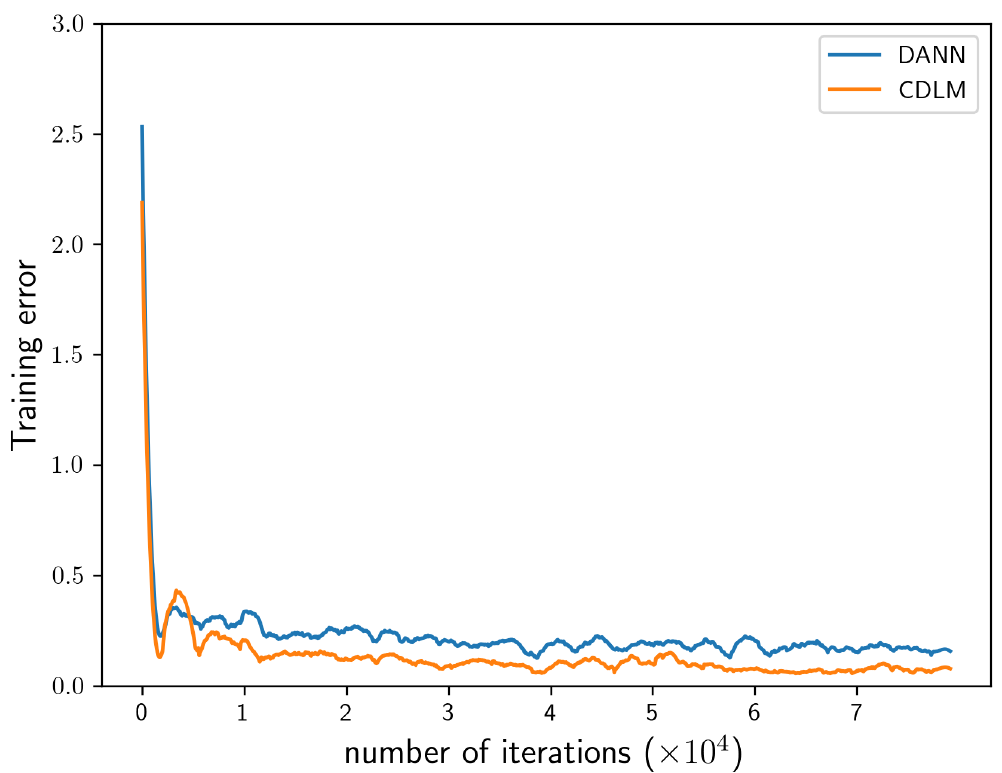}
  \caption{\small Convergence of CDLM compared with DANN.}
  \label{fig:ch4_convergence}
\end{figure}

\noindent
\textbf{Convergence:} We also conduct the convergence experiment using the training error on task MNIST-USPS to evaluate our model. Fig.~\ref{fig:ch4_convergence} shows that our model has better convergence than DANN, though there are some oscillations at the beginning of the training. The error of CDLM quickly drops lower than DANN after about 5000 iterations, indicating its better adaptation performance. This is consistent with the adaptation performance given in Table~\ref{table:ch4_accuracy}.

\section{Conclusion} \label{sec:conclusion}

In this paper, we have presented a novel variational cross-domain transfer learning model based on cross-domain latent modulation of deep representations from different domains. A shared transfer latent space is introduced, and a new reparameterization transformation is developed to implement the cross-domain modulations. Evaluations carried out in unsupervised domain adaptation and image translation tasks demonstrate our model's competitive performance. Its effectiveness is also clearly shown in visual assessment of the adapted images, as well as in the alignment of the latent information as revealed by $\mathcal{A}$-distance evaluation, as well as by visualization using t-SNE. Overall, competitive performance has been achieved by our model despite its relative simplicity compared with the-state-of-the-art methods. 

We have considered only the one-on-one transfer scenario for both domain adaptation and image translation tasks. Our variational cross-domain mechanism, however, seems plausible for an extension to deal with multi-domain transfer learning tasks. An interesting application would be to translate images to multiple target styles through the cross-modulation of multiple style priors, which may, for example, enable medical images to be generated across multiple modalities. For future work, we intend to further improve our variational transfer learning framework and use it for heterogeneous, multi-domain transfer tasks.   


\bibliography{cdlm}
\end{document}